\documentclass[sn-mathphys,Numbered]{sn-jnl}

\usepackage{graphicx}%
\usepackage{multirow}%
\usepackage{amsmath,amssymb,amsfonts}%
\usepackage{amsthm}%
\usepackage{mathrsfs}%
\usepackage[title]{appendix}%
\usepackage{xcolor}%
\usepackage{textcomp}%
\usepackage{manyfoot}%
\usepackage{booktabs}%
\usepackage{algorithm}%
\usepackage{algorithmicx}%
\usepackage{algpseudocode}%
\usepackage{listings}%
\usepackage{subcaption}
\usepackage{bbm}
\usepackage[capitalize]{cleveref}

\usepackage{color}
\definecolor{blue}{rgb}{0.0,0.0,1.0}

\def\eg{\emph{e.g.}} 
\def\ie{\emph{i.e.}} 
\def\etal{\emph{et al.}}

\theoremstyle{thmstyleone}%
\theoremstyle{thmstyletwo}%

\theoremstyle{thmstylethree}%

\begin{document}

\title[Article Title]{Towards Robust Semantic Segmentation against Patch-based Attack via Attention Refinement}


\author[1,2]{\fnm{Zheng} \sur{Yuan}}\email{zheng.yuan@vipl.ict.ac.cn}
\author*[1,2]{\fnm{Jie} \sur{Zhang}}\email{zhangjie@ict.ac.cn}
\author[1,2]{\fnm{Yude} \sur{Wang}}\email{yude.wang@vipl.ict.ac.cn}
\author[1,2]{\fnm{Shiguang} \sur{Shan}}\email{sgshan@ict.ac.cn}
\author[1,2]{\fnm{Xilin} \sur{Chen}}\email{xlchen@ict.ac.cn}

\affil[1]{\orgdiv{Key Laboratory of Intelligent Information Processing}, \orgname{Institute of Computing Technology, Chinese Academy of Sciences}, \orgaddress{\city{Beijing}, \postcode{100190}, \country{China}}}

\affil[2]{\orgname{University of Chinese Academy of Sciences}, \orgaddress{\city{Beijing}, \postcode{100049}, \country{China}}}


\abstract{The attention mechanism has been proven effective on various visual tasks in recent years. In the semantic segmentation task, the attention mechanism is applied in various methods, including the case of both Convolution Neural Networks (CNN) and Vision Transformer (ViT) as backbones.
However, we observe that the attention mechanism is vulnerable to patch-based adversarial attacks.
Through the analysis of the effective receptive field, we attribute it to the fact that the wide receptive field brought by global attention may lead to the spread of the adversarial patch.
To address this issue, in this paper, we propose a Robust Attention Mechanism (RAM) to improve the robustness of the semantic segmentation model, which can notably relieve the vulnerability against patch-based attacks. Compared to the vallina attention mechanism, RAM introduces two novel modules called Max Attention Suppression and Random Attention Dropout, both of which aim to refine the attention matrix and limit the influence of a single adversarial patch on the semantic segmentation results of other positions.
Extensive experiments demonstrate the effectiveness of our RAM to improve the robustness of semantic segmentation models against various patch-based attack methods under different attack settings.}




\maketitle

\section{Introduction}\label{sec:intro}

  In recent years, research on deep learning models develops rapidly, and a large number of models are widely used in various tasks, such as image classification~\cite{he2016deep, szegedy2016rethinking, zoph2018learning, tan2021efficientnetv2}, semantic segmentation~\cite{chen2015semantic, chen2018deeplab, chen2017rethinking}, and face recognition~\cite{liu2017sphereface, wang2018cosface, deng2019arcface}. With the prevalent application of the models, the security of deep learning models has become an increasingly concerning issue. For example, putting an adversarial patch on the face can deceive the face recognition model to identify it as another specific person~\cite{xiao2021improving, wei2022adversarial}, which could lead to serious security threats in access control systems and face payment scenarios.
  However, in the current research on model robustness, the majority of attention is focused on classification models~\cite{madry2018towards, serrurier2021achieving, wang2019on}, while other tasks such as semantic segmentation~\cite{kamann2020benchmarking, kamann2020increasing} and object detection~\cite{xie2017adversarial, michaelis2019benchmarking} have received relatively less attention.
  
  In this work, we focus on the robustness of semantic segmentation models against patch-based adversarial attacks. In the application of semantic segmentation models, compared to pixel-wise attack (\ie, the adversary can add adversarial perturbations to any pixel of the image), patch-based attack (\ie, the adversary can only modify a small patch of the image) is a more practical attack setting. For example, in the scenario of autonomous driving, an adversary can place an elaborately designed patch on a roadside sign, causing the semantic segmentation model deployed in the car to ignore the sign and make some rule-breaking decisions.

\begin{figure*}[]
  \centering
  \includegraphics[width=\linewidth]{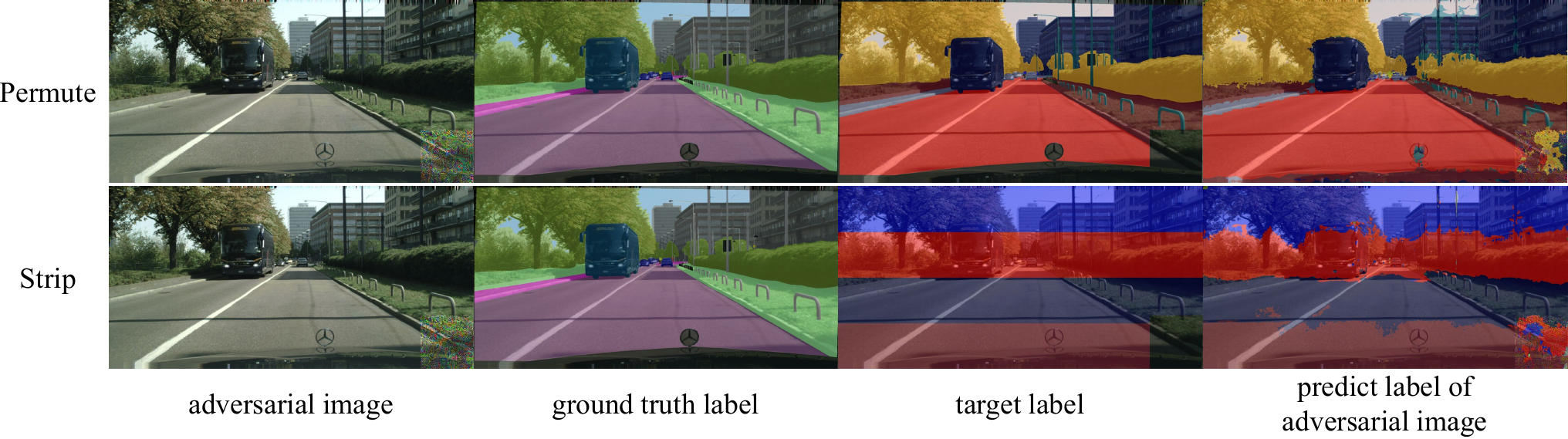}
  \caption{A visualization example of the targeted attack setting of Permute and Strip. \textbf{Permute} refers to the setting that replaces the labels in the ground truth segmentation with labels that do not appear in them one by one. \textbf{Strip} refers to the setting that chooses a strip pattern consisting of randomly selected labels as the target segmentation label.}
  \label{fig:att_setting}
\end{figure*}

  We first analyze the relationship between the effective receptive field of semantic segmentation models and their robustness against patch-based attacks.
  Specifically, we study the robustness under two targeted attack settings, \ie, Permute and Strip, as explained in detail in~\cref{sec:setting}. A visualization example of these settings is presented in~\cref{fig:att_setting}.
  Our analysis, as shown in~\cref{tab:receptive}, reveals that semantic segmentation models with wider receptive fields are more vulnerable to patch-based adversarial attacks.
  We attribute it mainly to the attention mechanism in the semantic segmentation model.
  In the task of semantic segmentation, the attention mechanism is a double-edged sword. It enables the model to predict the semantic labels of each location by incorporating contextual information from the local neighborhood to the global image. However, when the model is under patch-based attack, the adversary can exploit the attention matrix to increase the importance of a local dirty patch to other positions, spreading its influence globally. As a result, the output of any position in the image may be impacted by the local dirty patch.

  Based on the above findings, we propose a Robust Attention Mechanism (RAM) to improve the robustness of attention-based semantic segmentation models against patch-based attacks. Our RAM achieves this by limiting the influence of a single patch on other positions of the image.
  Compared to the conventional self-attention mechanism, as shown in~\cref{fig:RAM}, RAM introduces two novel modules called Max Attention Suppression (MAS) and Random Attention Dropout (RAD). Specifically, the MAS module limits the upper bound of a single element in the attention matrix, thereby reducing the impact of the adversarial patch on other positions. The RAD module performs a dropout operation on the attention matrix obtained in the attention mechanism. With this module, only part of the patches is considered when using the attention matrix as the weight to update the feature at each position, so that the influence of the potential adversarial patch can be eliminated with a certain probability.

  Our proposed RAM is simple but effective, and can be applied to almost all attention-based semantic segmentation models, which include the methods with both CNN and ViT models as backbones. Extensive experiments demonstrate the effectiveness of our RAD to improve the robustness of semantic segmentation models against various patch-based attack methods under different attack settings. Specifically, on the ADE20K~\cite{zhou2017scene} dataset, our RAM reduces the mIoU between the segmentation results of adversarial images and the target labels determined by the adversary up to about 20\% and 10\% averagely on global attention-based and local attention-based models, respectively. 

  We summarize our main contributions as follows:

  
  1. Through the analysis of the effective receptive field, we conclude that the vulnerability of models against patch-based attacks is attributed to the attention mechanism used in semantic segmentation models.

  2. We propose a Robust Attention Mechanism (RAM), which introduces two novel modules called Max Attention Suppression and Random Attention Dropout, and notably improves the robustness against patch-based attacks.

\begin{table*}[!t]
  \caption{Visualization of effective receptive fields of different semantic segmentation models (zoom-in for better quality) and corresponding model robustness against two patch-based targeted attack settings, \ie, Permute and Strip. The details of the settings are illustrated in~\cref{sec:setting} and a visualization example is shown in~\cref{fig:att_setting}. The mean Intersection-over-Union (mIoU) between the segmentation results of the adversarial images and the target labels determined by the adversary is utilized to evaluate the robustness of the model. The lower mIoU means a more robust model. Details about the visualization of the receptive field are illustrated in \cref{sec:receptive}.}
  \centering
  \resizebox{\textwidth}{!}{
    \begin{tabular}{c|c|c|c}
      \toprule
      \multirow{2}{*}{\begin{tabular}[c]{@{}c@{}}Model\\ (Head/Backbone)\end{tabular}} & No Attention & Local Attention & Global Attention \\ \cmidrule{2-4} 
      & UPerNet~\cite{xiao2018unified}/ConvNeXt~\cite{liu2022a}
      & SeMask~\cite{jain2021semask}/Swin~\cite{liu2021swin}
      & Segmenter~\cite{strudel2021segmenter}/ViT~\cite{dosovitskiy2021an} \\
      \midrule
      Receptive Field 
      &\raisebox{-.5\height}{\includegraphics[width=0.3\linewidth]{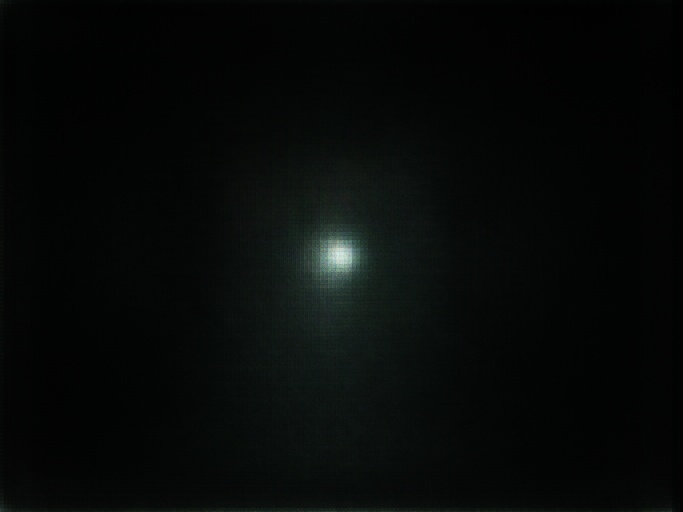}}
      &\raisebox{-.5\height}{\includegraphics[width=0.3\linewidth]{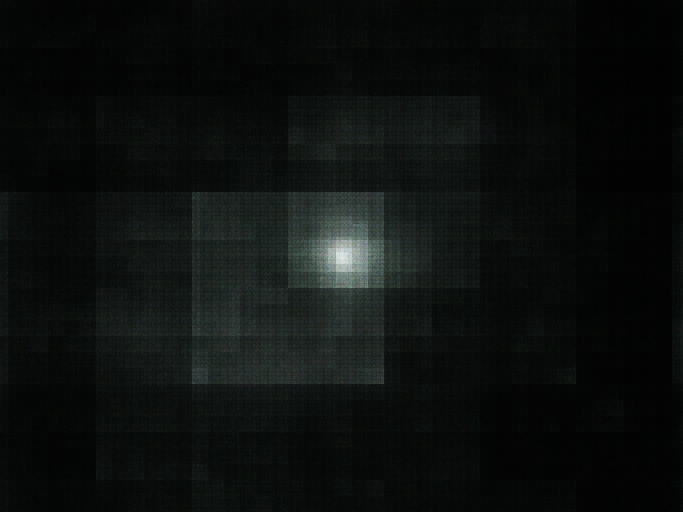}}
      &\raisebox{-.5\height}{\includegraphics[width=0.3\linewidth]{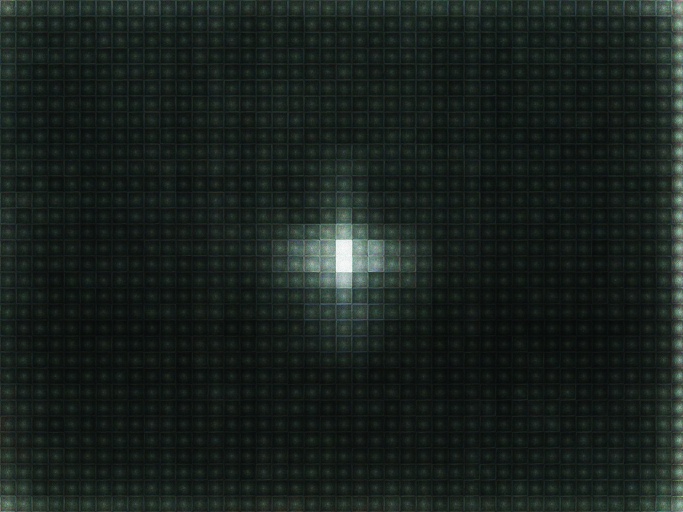}} \\
      \midrule
      Permute (mIoU/\%) & 5.86 & 29.25 & 74.57 \\
      Strip (mIoU/\%)  & 4.67 & 26.11 & 64.21 \\
      \bottomrule
    \end{tabular}
  }
  \label{tab:receptive}
\end{table*}

\section{Related Work}\label{sec:related}

  Our work focuses on the robustness of semantic segmentation models with an attention mechanism against patch-based attacks. Since the attention mechanism is widely used in ViT-based models, we briefly introduce the related works about semantic segmentation models, patch-based adversarial attacks and the robustness of ViT models, respectively.

\subsection{Semantic Segmentation}

  Early semantic segmentation methods use CNN models as backbones to extract features. FCN~\cite{long2015fully} is the first to propose fully convolutional networks, which can adapt to any resolution of input images. Since CNN extracts features by stacking multiple layers of convolution, there is a contradiction between preserving the receptive field and preserving resolution in earlier works. DeepLab~\cite{chen2015semantic} proposes a dilated convolution layer. It enlarges the receptive field of networks without introducing additional parameters. And some works~\cite{chen2018deeplab, zhao2017pyramid} adopt multi-scale feature aggregation modules to guarantee both the large-scale receptive field and high-resolution feature map. Self-attention~\cite{wang2018nonlocal} is another important module that introduces global receptive filed for each pixel, and its subsequent works~\cite{yuan2018ocnet, huang2019ccnet, li2019expectation} adopt various methods to reduce the complexity. Based on self-attention, vision transformer~\cite{dosovitskiy2021an} has become the mainstream model in various computer vision tasks, and its variants in semantic segmentation~\cite{zheng2021rethinking, strudel2021segmenter, xie2021segformer, jain2021semask, cheng2021perpixel, cheng2022masked} also achieve promising results. Recently, SegViT v2~\cite{segvit2024zhang} introduces a novel Attention-to-Mask (ATM) module to design a lightweight decoder effective for plain ViT.

\subsection{Patch-based Adversarial Attack}
  Patch-based adversarial attack means that the adversary can add noise to a patch area in the input image, thereby misleading the model to get the wrong output. Brown \etal~\cite{brown2017adversarial} first proposes to create universal and robust targeted adversarial image patches in the real world. Subsequent work proposes corresponding patch-based attack methods for various visual tasks, such as image classification~\cite{karmon2018lavan, yang2020patchattack}, object detection~\cite{liu2019dpatch, lee2019on, hu2021naturalistic}, autonomous driving~\cite{sato2021dirty} and face recognition~\cite{xiao2021improving, wei2022adversarial}.

  In the task of semantic segmentation, Nakka \etal~\cite{nakka2020indirect} introduces an indirect local attack where perturbations are confined to a small image region that does not overlap with the area that aims to fool. IPatch~\cite{mirsky2021ipatch} is then proposed to utilize remote adversarial patches to change the semantics of locations far from the patch. Nesti~\etal~\cite{nesti2022evaluating} extends the EOT paradigm~\cite{athalye2018synthesizing} to semantic segmentation models and evaluates the robustness against both digital and real-world adversarial patches.

  In addition, some works~\cite{fu2022patch, lovisotto2022give} propose patch-based attacks on the ViT model, but existing works are mainly focused on image classification models.
\begin{table*}[!t]
  \centering
    \caption{The robustness of semantic segmentation models against two patch-based targeted attack settings. The results in the table are the mIoU between the segmentation results of adversarial images and the target labels determined by the adversary. The lower mIoU means the better robustness of the model. Details about the attack settings are illustrated in \cref{sec:setting}.}
    \resizebox{\textwidth}{!}{
      \begin{tabular}{c|c|c|c|c|c}
        \toprule
        \multirow{2}{*}{\begin{tabular}[c]{@{}c@{}}Model\\ (Head/Backbone)\end{tabular}} & \multicolumn{2}{c|}{No Attention} & Local Attention & \multicolumn{2}{c}{Global Attention} \\
        \cmidrule{2-6} 
        & FCN~\cite{long2015fully}/R101~\cite{he2016deep} & FPN~\cite{kirillov2019panoptic}/PoolFormer~\cite{yu2022metaformer} & UPerNet~\cite{xiao2018unified}/Swin~\cite{liu2021swin} & Nonlocal~\cite{wang2018nonlocal}/R50~\cite{he2016deep} & UPerNet~\cite{xiao2018unified}/DeiT~\cite{touvron2021training} \\
        \midrule
        Permute$\downarrow$ (mIoU/\%) & 7.01 & 9.57 & 20.81 & 40.42 & 79.16 \\
        Strip$\downarrow$ (mIoU/\%) & 7.05 & 9.47 & 16.52 & 38.66 & 85.11 \\
        \bottomrule
      \end{tabular}
    }
  \label{tab:robustness}
\end{table*}
  
\subsection{Robustness of Vision Transformer}
  With the increasing popularity of the ViT model, several works~\cite{benz2021adversarial, wang2022can, bai2021are, mahmood2021on, bhojanapalli2021understanding} compare the robustness of the ViT model and the CNN model against adversarial examples, and find that the ViT model is more robust in most scenarios.
  Some works~\cite{bai2022improving, shao2022on} further study the robustness of the ViT model from the perspective of the frequency analysis, and find that the ViT models are less effective in capturing the high-frequency components of images than CNN models. 
  Adversarial training~\cite{debenedetti2022a, wu2022towards, rando2022exploring, herrmann2022pyramid} and certified patch defenses~\cite{salman2022certified, chen2022towards, huang2021zero} are also applied to ViT models to improve the robustness against adversarial examples.
  In addition, some works modify the structure of ViT model to improve the model's robustness. Mao \etal~\cite{mao2022towards} conducts systematic evaluation on components of ViTs in terms of their impact on robustness to adversarial examples and proposes a robust Vision Transformer model. Gu \etal~\cite{gu2022are} proposes a simple Smoothed Attention to discourage the attention mechanism to a single patch.
  However, almost all existing works are aimed at improving the robustness of classification models, and there are few works on how to improve the robustness of ViT-based semantic segmentation models against patch-based adversarial attacks.


\section{Analyses of Receptive Field and Robustness}
\label{sec:receptive}
  In this section, we analyze the relationship between the effective receptive field of the semantic segmentation models and their robustness against patch-based attacks.
  Our findings suggest that semantic segmentation models with wider receptive fields are more vulnerable to patch-based adversarial attacks. We attribute this phenomenon primarily to the presence of attention mechanisms within the semantic segmentation model.

  To explore this relationship further, we first introduce a visualization method of the effective receptive field. And then we show the impact of the attention mechanism by visualizing the effective receptive fields of different models and comparing the robustness of different models against patch-based attacks.

\subsection{Receptive Field}
  Suppose the input image is $\mathbf{X} \in \mathbb{R}^{H\times W\times 3}$, the semantic segmentation model is $f(\cdot)$, and the segmentation result of the model for the input $\mathbf{X}$ is denoted by $\mathbf{Y}=f(\mathbf{X})\in \mathbb{R}^{H\times W\times C}$, where $H$ and $W$ are the height and width of the input image, and $C$ is the number of categories for image segmentation.
  To visualize the effective receptive field of the model, we follow the method proposed in~\cite{luo2016understanding}. 
  Specifically, we back-propagate the gradient of the central output unit to the input as the receptive field of the model:
  \begin{equation}
    \sum_{c=0}^{C-1}\frac{\partial \mathbf{Y}_{H/2,W/2,c}}{\partial \mathbf{X}},
  \end{equation}
  where $\mathbf{Y}_{ijc}$ is the predicted probability of pixel belonging to class $c$ at position $(i, j)$. We randomly select 1000 images from the validation set in the ADE20K dataset~\cite{zhou2017scene} as the input and visualize their average receptive field.

\subsection{Analysis}

  The visualization of the effective receptive field of different models and the corresponding robustness against patch-based attacks are shown in~\cref{tab:receptive}.
  In the experiments, we evaluate the robustness of models against two different attack targets, named Permute and Strip, respectively. A visualization example of the two targeted attack settings is shown in~\cref{fig:att_setting} and the details of attack settings are described in~\cref{sec:setting}. For the patch-based attack, we use the method of DAG~\cite{xie2017adversarial} by adding an adversarial patch with the size of $150\times 150$ in the lower right corner of the image. We use the mIoU between the segmentation results of the adversarial examples and the target labels determined by the adversary as the metric to evaluate the robustness of the semantic segmentation model against these attacks. Lower mIoU values indicate the better robustness of the model.

  From~\cref{tab:receptive}, we can see that the effective receptive field of the model is relatively small when using ConvNeXt~\cite{liu2022a}, a model without attention mechanism, as the backbone of semantic segmentation model (left column in~\cref{tab:receptive}).
  Additionally, the model appears to be more robust to patch-based attacks, as indicated by the small mIoU values between the segmentation predictions of adversarial images and the target segmentation label under both Permute and Strip attack settings.
  In contrast, the effective receptive field of the semantic segmentation model almost covers the whole images when using ViT~\cite{dosovitskiy2021an} as the backbone (right column in~\cref{tab:receptive}), which has a global attention mechanism. At the same time, the model with self-attention is also significantly more vulnerable to patch-based attacks.
  Swin Transformer~\cite{liu2021swin} utilizes the shifted window based self-attention to replace the global attention mechanism used in ViT. The local attention mechanism in Swin Transformer leads to a moderate range of receptive field (middle column in~\cref{tab:receptive}), and the corresponding model robustness is also in the middle level between the model with traditional convolution and global attention mechanism.
    
  More results about the robustness of different models are shown in~\cref{tab:robustness}. It indicates that the attention mechanism plays a crucial role in enhancing the robustness of the semantic segmentation models against patch-based attacks, regardless of whether a ViT-based backbone is used or not. An alternative model, PoolFormer~\cite{yu2022metaformer}, replaces the attention module in ViT with a spatial pooling operator, resulting in a significant improvement in the robustness of the semantic segmentation model. On the contrary, Nonlocal~\cite{wang2018nonlocal} method uses the CNN model as the backbone and performs global attention operation on the extracted features, yet the resulting model is still vulnerable to patch-based attacks.
  
  In the task of semantic segmentation, the attention mechanism is a double-edged sword.
  The attention mechanism promotes the model to well predict the semantic labels of each pixel by enriching the contextual information from both local and global ranges. However, in the case of a patch-based attack, the attention mechanism can be exploited by adversaries to elevate the significance of a dirty patch and propagate its influence to other positions via the attention matrix. This can result in the global spread of the patch's impact, potentially affecting the output of any position in the image.

  It is worth mentioning that Mao \etal~\cite{mao2022towards} find that compared with the global attention mechanism in ViT~\cite{dosovitskiy2021an}, the local attention used in Swin~\cite{liu2021swin} damages the robustness of classification models against pixel-wise attacks. It is opposite to the conclusion we draw for the robustness of semantic segmentation models against patch-based attacks in our work. It reflects the unique value of the robustness research on semantic segmentation models under the patch-based attack settings, and highlights the contributions of our work.
  We attribute the opposite conclusions to the combined effects of different attack types (pixel-wise vs. patch-based) and different task models (classification vs. segmentation).
  In classification tasks, under pixel-wise attacks, the entire image is subjected to noise interference. To make a correct classification, models may not need the information from the entire image, but can rely on distinctive local content for decision-making. When the critical discriminative region is disrupted, global attention methods, with their larger receptive fields, can still generate the correct label by relying on information from other locations. 
  For segmentation tasks during patch-based attacks, the scenario differs as segmentation requires classifying all pixels in an image. When an image suffers a patch-based attack, the adversarial perturbation appears only in a specific part of the image. Global attention might spread the localized disturbance across the entire image, potentially affecting the accuracy of segmentation results in more areas.
  

\section{Robust Attention Mechanism}
\label{sec:model}

\begin{figure}[!t]
  \centering
  \includegraphics[width=0.7\linewidth]{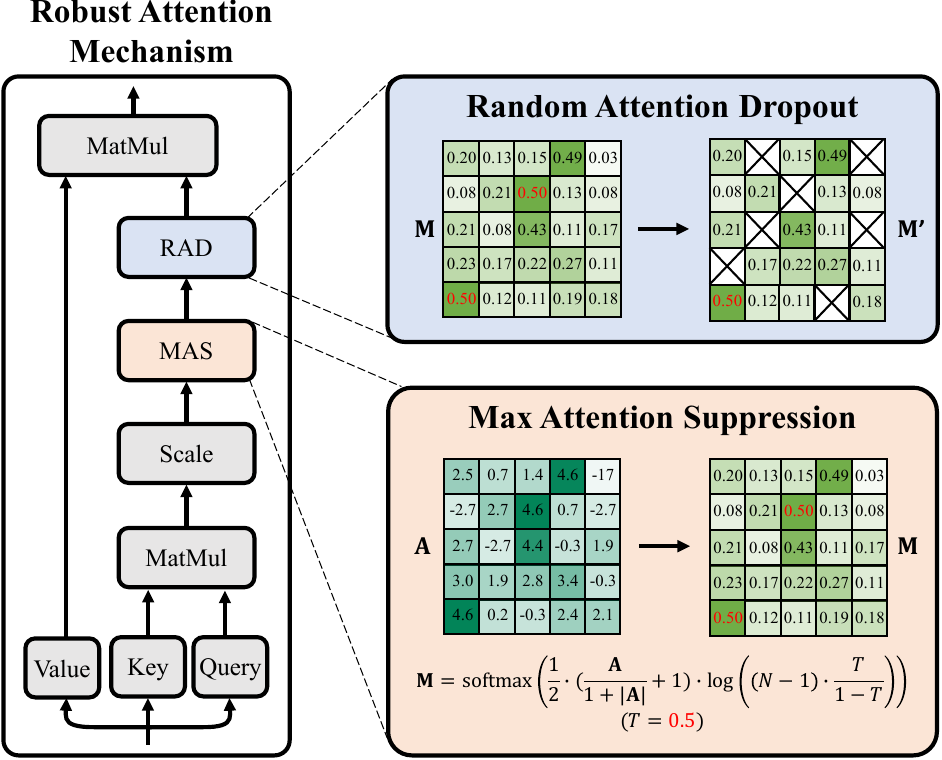}
  \caption{The structure of our proposed Robust Attention Mechanism (RAM), which introduces two novel modules called Max Attention Suppression (MAS) and Random Attention Dropout (RAD).}
  \label{fig:RAM}
\end{figure}

  In order to increase the robustness of the semantic segmentation model with attention-based backbones against patch-based attacks, we propose the Robust Attention Mechanism (RAM). As shown in~\cref{fig:RAM}, compared to the vallina attention mechanism, our RAM introduces two additional modules, \ie, Max Attention Suppression and Random Attention Dropout, which will be introduced below, respectively.

\subsection{Vallina Attention Mechanism}
  The earliest application of the attention mechanism in semantic segmentation tasks appears in Nonlocal~\cite{wang2018nonlocal}. With the rapid development of ViT in recent years, the attention mechanism has become a mainstream and indispensable module. We first introduce the general attention mechanism. Suppose the image feature is $\mathbf{x}\in \mathbb{R}^{N\times C}$, where $N$ is the number of positions and $C$ is the channel of the feature.
  In the attention mechanism, the input feature $\mathbf{x}$ is first linearly transformed into three features named as key $\mathbf{K} \in \mathbb{R}^{N\times d_k}$, value $\mathbf{V}\in \mathbb{R}^{N\times d_v}$ and query $\mathbf{Q}\in \mathbb{R}^{N\times d_q}$, respectively. Then the matrix multiplication is performed on $\mathbf{K}$ and $\mathbf{Q}$, followed by scaling to obtain the attention matrix $\mathbf{A}$, which represents the relationship between the features for each pair of positions. Finally, the attention matrix is normalized by the softmax function and multiplied with the feature of value $\mathbf{V}$ to get the out feature $\mathbf{y}$. The whole process can be formulated as follows:
  \begin{gather}
    \mathbf{K} = \mathbf{x W_K^\top},\quad \mathbf{Q} = \mathbf{x W_Q^\top}, \quad \mathbf{V}=\mathbf{x W_V^\top}, \\
    \mathbf{A} = \frac{\mathbf{QK^\top}}{\sqrt{d_k}}, \quad \mathbf{M} = \text{softmax}(\mathbf{A}), \quad \mathbf{y}=\mathbf{MV},
  \end{gather}
  where $\mathbf{W_K} \in \mathbb{R}^{d_k \times C}$, $\mathbf{W_V} \in \mathbb{R}^{d_v \times C}$, and $\mathbf{W_Q} \in \mathbb{R}^{d_q \times C}$ are the learnable model parameters.

\subsection{Max Attention Suppression}
\label{sec:mas}
  In order to reduce the influence of the potential local dirty patch on the overall output of the segmentation model, we propose Max Attention Suppresion (MAS) module to limit the influence of a single patch on the feature of other locations. Specifically, we transform the attention matrix $\mathbf{A}$ as follows:
  \begin{gather}
    \mathbf{A'} = \frac{1}{2}\cdot\left(\frac{\mathbf{A}}{(1+|\mathbf{A}|)}+1\right)\cdot \log \left((N-1)\cdot\frac{T}{1-T}\right), \\
    \mathbf{M} = \text{softmax}(\mathbf{A'}) \label{equ:m},
  \end{gather} 
  where $T\in [0, 1]$ is a threshold to control the maximum of elements in the attention matrix.

  After the above transformation, the MAS module guarantees that the maximum element in the attention matrix normalized by softmax function $\mathbf{M}$ does not exceed $T$, \ie,
  \begin{equation}
    \label{equ:mas}
    \max_{i,j} \mathbf{M}_{ij} \leq T.
  \end{equation}

\begin{proof}

   For the attention matrix $\mathbf{A}$, since it is obtained from the matrix multiplication of the feature key $\mathbf{K}$ and query $\mathbf{Q}$, its range can be any real number, \ie,
   \begin{equation}
    \mathbf{A} \in \mathbb{R}^{N\times N},
   \end{equation} 
   where $N$ is the number of feature positions.
   Since the value range of the function $y=\frac{x}{1+|x|}$ is $[-1, 1]$, the corresponding value range of $\mathbf{A'}$ is $\left[0, \log \left((N-1)\cdot \frac{T}{1-T}\right)\right]$, \ie, 
   \begin{equation}
    \mathbf{A'} \in \left[0, \log \left((N-1)\cdot \frac{T}{1-T}\right)\right]^{N\times N}.
   \end{equation}
   Then we normalize the transformed attention matrix $\mathbf{A'}$ by the softmax function along the rows, \ie,
   \begin{equation}
     \label{equ:m_ik}
     \mathbf{M}_{ik}=\frac{\exp{(\mathbf{A'}_{ik})}}{\exp{(\mathbf{A'}_{ik})}+\sum_{j\neq k}\exp{(\mathbf{A'}_{ij})}}.
   \end{equation}
   Since the softmax function is monotonically increasing, so 
   \begin{equation}
     \forall i, \quad \arg \max_j \mathbf{A'}_{ij} = \arg \max_j \mathbf{M}_{ij}.
   \end{equation}
   Without loss of generality, we assume that among the elements in the $i$-th row of $\mathbf{A'}$, the index of the largest element is $k$, \ie, $\arg \max_j \mathbf{A'}_{ij}=k$.
 
   Let's examine the largest element in the $i$-th row of matrix $\mathbf{M}$, \ie, $\mathbf{M}_{ik}$.
   When we fix all $\mathbf{A'}_{ij}, \forall j\neq k$, the function in \cref{equ:m_ik} is monotonically increasing with $\mathbf{A'}_{ik}$.
   And when we fix $\mathbf{A'}_{ik}$, the function in \cref{equ:m_ik} is monotonically decreasing with $\mathbf{A'}_{ij}, \forall j\neq k$.
   Since $\mathbf{A'}_{ij}\in\left[0, \log \left((N-1)\cdot \frac{T}{1-T}\right)\right]$, we have
   \begin{align}
    \mathbf{M}_{ik}&= \frac{\exp{(\mathbf{A'}_{ik})}}{\exp{(\mathbf{A'}_{ik})}+\sum_{j\neq k}\exp{(\mathbf{A'}_{ij})}} \\
     &\leq \frac{\exp\left(\log \left((N-1)\cdot \frac{T}{1-T}\right)\right)}{\exp\left(\log \left((N-1)\cdot \frac{T}{1-T}\right)\right) + (N-1)\cdot\exp{(0)}}\\
     &=\frac{(N-1)\cdot \frac{T}{1-T}}{(N-1)\cdot \frac{T}{1-T} + (N-1)}\\
     &=T.
   \end{align}
   In summary, we have proved that 
   \begin{equation}
     \max_{i,j} \mathbf{M}_{ij} \leq T.
   \end{equation}
\end{proof}

  With the MAS module, we limit the maximum value of the elements in the attention matrix, thus ensuring that even when the model is under the patch-based attack, the negative impact of the dirty patch on other locations will not be particularly large.

  Although imposing additional restrictions on the elements in the attention matrix may lead to a reduction in model performance, our subsequent experiments demonstrate that the decrease in mIoU for semantic segmentation results on clean images is almost negligible. In contrast, the improvement of model robustness against patch-based attacks is quite significant.

\subsection{Random Attention Dropout}
  To further improve the robustness of the attention module against potential patch-based attacks, we propose a Random Attention Dropout (RAD) module to conduct dropout operation~\cite{srivastava2014dropout} on the attention matrix with a certain probability:
  \begin{equation}
    \mathbf{M'} = \text{dropout}(\mathbf{M},p),\quad \mathbf{y=M'V},
  \end{equation}
  where $\text{dropout}(\mathbf{M},p)$ represents randomly dropping elements of $\mathbf{M}$ with the probability of $p$.

  With this module, when the model is attacked by patch-based adversarial examples, RAD can drop the dirty patch with a certain probability to directly eliminate the negative impact of the adversarial patch, thereby improving the robustness of the model against patch-based attacks.
  It should be noted that unlike the commonly used dropout operation, which is only applied during the training stage, our RAD is also retained at the evaluation stage (\ie, the scenario where the model is attacked) to ensure that the influence of the dirty patch may be eliminated with a certain probability.

\subsection{Discussion}
  In this section, we will discuss in detail the differences between our work and the related study~\cite{gu2022are}.
  Firstly, our work and~\cite{gu2022are} focus robustness analysis on different tasks: \cite{gu2022are} investi-gates the robustness of attention mechanisms in classification models against patch-based perturbations. Our research, however, studies the robustness of attention mechanisms in semantic segmentation models to patch-based perturbations. Given the distinct tasks of segmentation and classification models, patch-based attacks on segmentation tasks have unique characteristics. Semantic segmentation models are designed to provide segmen-tation results for every pixel within an entire image, where a local dirty patch could sig-nificantly influence the segmentation results through the attention mechanism, spreading its negative effect throughout the image. In contrast, image classification results may rely on features from key areas. Therefore, in image classification tasks, local dirty patches may not have a particularly significant impact on the overall classification result.
	Also, our work differs from~\cite{gu2022are} in methodology: \cite{gu2022are} suggests replacing the original attention with smoothed attention via temperature scaling in the softmax operation. Conversely, our method reduces the potential impact of dirty patches on other areas by compressing the value range of the attention map through max attention suppression and mitigates the negative effects of dirty patches by randomly dropping some patches using random attention dropout.

\section{Experiment}

\label{sec:exper}
  In this section, we will show the effect of our proposed RAM method to improve the robustness of semantic segmentation models against patch-based attacks through extensive experiments. 

\subsection{Experimental Setting}
\label{sec:setting}
  \textbf{Dataset.} We mainly use the ADE20K dataset~\cite{zhou2017scene} to conduct the experiments. The training and validation sets contain 20,210 and 2,000 images, respectively, with a total of 150 categories of semantic labels. We use the training set to train models, and use the validation set to evaluate the performance of the model and the robustness against patch-based attacks. We also conduct several experiments on VOC2012~\cite{everingham2015the} and Cityscapes~\cite{cordts2016the} datasets to demonstrate the effectiveness of our proposed RAM method.
 
  \textbf{Models.} We use various semantic segmentation models to verify the effectiveness of our proposed method. For CNN-based models, we use Nonlocal~\cite{wang2018nonlocal}/R50~\cite{he2016deep}, which with the global attention mechanism. For ViT-based models, we use both models with the global attention mechanism (\eg, SegFormer~\cite{xie2021segformer}/MiT~\cite{xie2021segformer}, Segmenter~\cite{strudel2021segmenter}/ViT~\cite{dosovitskiy2021an} and UPerNet~\cite{xiao2018unified}/DeiT~\cite{touvron2021training}) and models with the local attention mechanism (\eg, SeMask~\cite{jain2021semask}/Swin~\cite{liu2021swin} and UPerNet~\cite{xiao2018unified}/Swin~\cite{liu2021swin}).
  For the training of models, we refer to the settings in the mmsegmentation\footnote{\url{https://github.com/open-mmlab/mmsegmentation}} framework.

  \textbf{Comparison methods.} Due to the lack of research on improving the robustness of semantic segmentation models against patch-based attacks, we transfer existing methods for improving robustness against patch-based attacks in classification models to the semantic segmentation task and compare them with our RAM method. Specifically, Mao~\cite{mao2022towards} proposes to multiply the attention matrix with a learnable parameter to improve the robustness. Gu \etal~\cite{gu2022are} proposes to smooth the attention matrix by dividing it with a preset temperature. In the experiments, we refer to these two methods as Learnable and Temperature, respectively.

  \textbf{Attack methods.} We use totally ten attack methods (\ie, PGD~\cite{madry2018towards}, DAG~\cite{xie2017adversarial}, IPatch~\cite{mirsky2021ipatch}, SSAP~\cite{nesti2022evaluating}, Patch-Fool~\cite{fu2022patch}, Attention-Fool~\cite{lovisotto2022give}, AutoAttack~\cite{croce2020reliable}, EOT~\cite{athalye2018synthesizing}, MaxVarDAG and MaxAttnDAG) to evaluate the robustness of model. The details of these methods are illustrated in~\cref{sec:att_method}. Unless mentioned, all patch-based attacks are conducted by adding a $150\times 150$ patch on the lower right corner of the clean image. We use the DAG~\cite{xie2017adversarial} method to conduct the experiments by default.

  \textbf{Attack settings.} In the experiments, we conduct attacks under both targeted and untargered settings. For the targeted attack setting, we set up two different kinds of attack targets, named Permute and Strip, respectively.
  Suppose the ground truth segmentation label corresponding to the input image $\mathbf{X}$ is $\mathbf{Y^g} \in \{0,1,\cdots, C-1\}^{H\times W}$ and the target segmentation label of the attack is $\mathbf{Y^t}\in \{0,1,\cdots, C-1\}^{H\times W}$. Suppose the set of categories that appears in $\mathbf{Y^g}$ is $S=\{y_1,...,y_n\}\subseteq \{0,1,\cdots, C-1\}$.
  
  Under the attack setting of Permute, we permute the set $S=\{y_1,...,y_n\}$ to a subset that belongs to the complement of $S$, \ie, $S^t=\{y_1^t,...,y_n^t\} \subseteq \{0,1,\cdots, C-1\} \backslash S$, as the target segmentation label of the attack $\mathbf{Y^t}$.
  Specifically, for all pixels in $\mathbf{Y^g}$ belonging to $y_c$, we set the corresponding attack target in $\mathbf{Y^t}$ as another random label $y_c^t$ that does not appear in $\mathbf{Y^g}$. The process can be formally expressed as:
  \begin{gather}
    S_c=\{(i,j)|\mathbf{Y^g}_{ij}=y_c\}, \quad \mathbf{Y^t}_{S_c}=y^t_c, \\
    y^t_c= \sigma (\{0,1,\cdots, C-1\}\backslash S),
  \end{gather}
  where $\sigma(S)$ represents randomly selecting an element from the set $S$.
  
  Under the attack setting of Strip, the target label of the attack $\mathbf{Y^t}$ is a strip pattern consisting of $N$ stripes, and the corresponding categories are also randomly selected from the categories that do not appear in $\mathbf{Y^g}$, which is formally expressed as:
  \begin{gather}
    S_n=\left\{(i,j)|i\in\{\frac{nH}{N}, \cdots, \frac{(n+1)H}{N}\}, j\in\{0,\cdots, W\}\right\}, \\
    \mathbf{Y^t}_{S_n}=y^t_n, \quad y^t_n = \sigma (\{0,1,\cdots, C-1\}\backslash S).
  \end{gather}
  
  A visualization example of the two targeted attack settings is shown in~\cref{fig:att_setting}. It should be noted that our reason for adopting targeted attack settings, in addition to the common untargeted attack scenarios, is to simulate more practical scenarios. In the context of untargeted attacks, the segmentation results of the perturbed images are significantly disrupted, making it easier for the model providers to visually inspect the model outputs and detect the model is under attack, thereby enabling the enactment of specific countermeasures. Consequently, we resort to the targeted attack and use target labels designated by the attackers themselves, producing adversarial examples (\eg, under the Permute setting) that, although bearing incorrect semantic category labels, remain shape-wise similar to the ground truth labels (as shown in~\cref{fig:att_setting}), \ie, the high similarity between edge contours of ground-truth and the adversarial labels, making them relatively harder for model providers to visually detect.

  \textbf{Metrics.} Following previous semantic segmentation works, we adopt mean Intersection-over-Union (mIoU) as the metric to evaluate the semantic segmentation model for clean images. Besides, to assess the robustness of models under two targeted attack settings, we compare the segmentation results of the adversarial examples and the target labels determined by the adversary, and calculate their mIoU as the metric. Here, the smaller the mIoU means the robustness of the model is better.
  For the untargeted attack setting, we use the metric of the mIoU between the segmentation results of the adversarial images and the ground truth labels to evaluate the robustness of the model.
  In order to fully demonstrate the effectiveness of our method, we also use the metric of pixel accuracy (pAcc) to evaluate the robustness of different models. Specifically, we calculate pAcc as follows:
  \begin{gather}
      pAcc = \frac{\sum_{i=0}^{H-1}\sum_{j=0}^{W-1} \mathbbm{1}\{\arg\max_c \mathbf{Y}_{ijc} =\mathbf{Y^g}_{ij} \} \odot (\mathbf{1_{H\times W}} - \mathbf{M_{mask}})}{H\cdot W - m^2},
  \end{gather}
  where $\mathbf{M_{mask}} \in \{0,1\}^{H\times W}$ is the onehot mask of the adversarial patch, $\mathbbm{1}\{\cdot\}$ refers to the indicator function, and $m$ is the side length of the patch.

\subsection{Ablation Study}

  We perform an ablation study on the proposed Robust Attention Module to verify the effects of Max Attention Suppression and Random Attention Dropout modules on improving the robustness of the semantic segmentation model. As shown in~\cref{tab:ablation}, compared with the baseline method, our proposed MAS and RAD modules can improve the robustness of the model, respectively. Especially for the MAS module, the target mIoU under attack drops from 49\% to 34\% under Permute setting, which demonstrates that MAS can significantly improve the robustness of semantic segmentation against patch-based attacks. When combining our proposed MAS and RAD together, the complete RAM can further improve the robustness of the model.

\begin{table}[]
  \centering
  \caption{Ablation study of our proposed RAM.
  The results in the table are the mIoU between the segmentation results of the adversarial images and the target labels determined by the adversary.
  }
  \begin{tabular}{cc|ccc}
    \toprule
    \multicolumn{2}{c|}{Setting} & {clean$\uparrow$} & {Permute$\downarrow$} & {Strip$\downarrow$} \\
    \cmidrule{1-2}MAS   & RAD   & (mIoU/\%) & (mIoU/\%) & (mIoU/\%) \\
    \midrule
    $\times$ & $\times$ & 48.25  & 49.32  & 40.01  \\
    $\times$ & $\surd$ & 48.37  & 46.77  & 37.83  \\
    $\surd$  & $\times$ & 48.10  & 34.11  & 23.36  \\
    $\surd$  & $\surd$ & \textbf{48.30} & \textbf{33.72} & \textbf{22.63} \\
    \bottomrule
  \end{tabular}
  \label{tab:ablation}
\end{table}

\subsection{The Impact of the Hyperparameter}
  In this section, we conduct the experiments to confirm the optimal hyperparameters in our proposed RAM, \eg, the threshold $T$ in Max Attention Suppression and the dropout rate $p$ in Random Attention Dropout.
  As $T$ decreases, the upper limit of the attention value associated with the dirty patch decreases, consequently mitigating its influence on other patches. However, when $T$ is too small, the compressed range of the attention map values $[0,T]$ becomes too limited, leading to a lack of differentiation between features of different patches, which could negatively affect the model's performance. Therefore, as shown in~\cref{tab:threshold}, as $T$ decreases, the results of our method initially improve but then deteriorate. Through ablation experiments, we have ultimately chosen $T=0.3$ as the hyperparameter.
  The parameter $p$ is the dropout rate if our proposed RAD method. The larger $p$ is, the greater the number of patches randomly dropped, increasing the probability of the dirty patch being randomly discarded, thereby reducing its impact on other clean patch locations. On the other hand, discarding more clean patches can affect the segmentation results. Hence, as shown in~\cref{tab:dropout}, as $p$ increases, the robustness of the model initially improves but then decreases. Ultimately, we select $p=0.5$ as the hyperparameter.

\begin{table}[]
  \centering
  \caption{The robustness of semantic segmentation models with our proposed RAM against two patch-based targeted attack settings under different thresholds $T$.
  The results in the table are the mIoU between the segmentation results of the adversarial images and the target labels determined by the adversary.
  }
    \begin{tabular}{c|ccccc}
    \toprule
    Threshold ($T$) & 0.2 & 0.3 & 0.4 & 0.5 & 0.6 \\ \midrule
    Permute$\downarrow$ (mIoU/\%)  & 35.12 & \textbf{34.11} & 35.63 & 37.63 & 37.69 \\
    Strip$\downarrow$ (mIoU/\%)   & 23.74 & \textbf{23.36} & 24.74 & 28.93 & 27.20    \\
    \bottomrule
    \end{tabular}

  \label{tab:threshold}
\end{table}

\begin{table}[]
  \centering
  \caption{The robustness of semantic segmentation models with our proposed RAM against two patch-based targeted attack settings under different dropout rates $p$.
  The results in the table are the mIoU between the segmentation results of the adversarial images and the target labels determined by the adversary.
  }
  \begin{tabular}{c|ccc}
  \toprule
  Dropout Rate ($p$) & 0.4 & 0.5 & 0.6 \\ \midrule
  Permute$\downarrow$ (mIoU/\%)  & 48.51 & \textbf{46.77} & 47.22 \\
  Strip$\downarrow$ (mIoU/\%)    & 40.05 & \textbf{37.83} & 38.30 \\
  \bottomrule
  \end{tabular}
  \label{tab:dropout}
\end{table}

\subsection{The Impact of the Size of Adversarial Patch}
\begin{table}[]
 \centering
 \caption{The robustness of semantic segmentation models against different sizes of patch-based attacks.
 The results in the table are the mIoU between the segmentation results of the adversarial images and the target labels determined by the adversary.
 }
   \begin{tabular}{c|cc|cc}
     \toprule
     \multirow{3}{*}{Patch Size} & \multicolumn{2}{c|}{SegFormer~\cite{xie2021segformer}/MiT~\cite{xie2021segformer}} & \multicolumn{2}{c}{RAM (ours)} \\ \cmidrule{2-5} 
       & Permute$\downarrow$ & Strip$\downarrow$ & Permute$\downarrow$ & Strip$\downarrow$ \\
       & (mIoU/\%) & (mIoU/\%) & (mIoU/\%) & (mIoU/\%) \\ \midrule
     100  & 31.69 & 18.24 & 13.65 & 7.25 \\
     125  & 40.96 & 29.88 & 24.33 & 13.65 \\
     150  & 49.32 & 40.01 & 33.72 & 22.69 \\
     175  & 55.55 & 48.92 & 42.74 & 32.55 \\
     \bottomrule
     \end{tabular}
 \label{tab:size}
\end{table}
  We conduct experiments to investigate the impact of the size of the adversarial patch on the robustness of the model. We vary the size of the adversarial patch from 100 to 175. As shown in ~\cref{tab:size}, as the size increases, mIoU under both attack settings gradually becomes larger, which indicates that the adversarial patch is more aggressive. 
  However, under different adversarial patch sizes, our proposed RAM method consistently reduces mIoU, and achieves about 10\%-20\% mIoU reduction under various experimental settings. This also shows that our proposed method has good generalization ability and can work in different scenarios.

\subsection{The Impact of the Location of Adversarial Patch}
\label{sec:location}

\begin{table}[]
  \caption{The robustness of semantic segmentation models against different locations of patch-based attacks. The results in the table are the mIoU between the segmentation results of the adversarial images and the target labels determined by the adversary. The lower mIoU means the better robustness of the model.
  }
  \begin{tabular}{c|cc|cc}
    \toprule
    \multirow{3}{*}{Patch Location} & \multicolumn{2}{c|}{SegFormer~\cite{xie2021segformer}/MiT~\cite{xie2021segformer}}                                  & \multicolumn{2}{c}{RAM (ours)}                                \\ \cmidrule{2-5} 
    & Permute$\downarrow$ & Strip$\downarrow$ & Permute$\downarrow$ & Strip$\downarrow$ \\
    & (mIoU/\%) & (mIoU/\%) & (mIoU/\%) & (mIoU/\%) \\ \midrule
    Lower Left  & 49.32 & 40.01 & 33.72 & 22.69 \\
    Lower Right & 48.68 & 40.12 & 34.53 & 22.84 \\
    Top Left    & 47.49 & 39.24 & 33.76 & 23.81 \\
    Top Right   & 46.59 & 38.82 & 34.28 & 23.06 \\
    Center      & 52.39 & 46.78 & 38.41 & 28.04 \\
    Random      & 50.28 & 42.76 & 36.58 & 27.28 \\
    \bottomrule
  \end{tabular}
  \label{tab:location}
\end{table}

  We also conduct experiments to investigate the impact of the locations of the adversarial patch on the robustness of the model.
  We place the adversarial patch at the four corners, at the center, and at randomly chosen locations of the image, respectively.
  As shown in ~\cref{tab:location}, regardless of how the location of the adversarial patch changes in the image, our proposed RAM method can always effectively improve the robustness of the model against patch-based attacks, which also well demonstrates the good generalization of our proposed method.

\subsection{Comparison with Existing Methods}
\subsubsection{Targeted Attack Setting}
\begin{table*}[]
  \centering
  \caption{The robustness of different semantic segmentation models against two patch-based targeted attack settings. \textbf{clean} represents the mIoU of the semantic segmentation results on the validation set, and \textbf{Permute} and \textbf{Strip} represent the mIoU between the segmentation results of the adversarial images and the target labels determined by the adversary according to different attack settings. The higher mIoU regards to \textbf{clean} means the higher segmentation accuracy of the model for clean images, and the lower mIoU regards to \textbf{Permute} and \textbf{Strip} means the better robustness of the model.
  }
  \resizebox{\linewidth}{!}{
    \begin{tabular}{c|ccccccccc|cccccc}
      \toprule
      \multirow{2}{*}{\begin{tabular}[c]{@{}c@{}}Model\\ (Head/Backbone)\end{tabular}} & \multicolumn{9}{c|}{Global Attention}                                                                                                                                         & \multicolumn{6}{c}{Local Attention}                                             \\ 
      \cmidrule{2-16} 
      & \multicolumn{3}{c|}{Nonlocal~\cite{wang2018nonlocal}/R50~\cite{he2016deep}}     & \multicolumn{3}{c|}{Segmenter~\cite{strudel2021segmenter}/ViT~\cite{dosovitskiy2021an}}           & \multicolumn{3}{c|}{UPerNet~\cite{xiao2018unified}/DeiT~\cite{touvron2021training}} & \multicolumn{3}{c|}{SeMask~\cite{jain2021semask}/Swin~\cite{liu2021swin}}             & \multicolumn{3}{c}{UPerNet~\cite{xiao2018unified}/Swin~\cite{liu2021swin}} \\
      \midrule 
      & clean$\uparrow$ & Permute$\downarrow$  & \multicolumn{1}{c|}{Strip$\downarrow$} & clean$\uparrow$  & Permute$\downarrow$ & \multicolumn{1}{c|}{Strip$\downarrow$} & clean$\uparrow$     & Permute$\downarrow$    & Strip$\downarrow$    & clean$\uparrow$ & Permute$\downarrow$ & \multicolumn{1}{c|}{Strip$\downarrow$} & clean$\uparrow$    & Permute$\downarrow$    & Strip$\downarrow$    \\
      & (mIoU/\%) & (mIoU/\%)  & \multicolumn{1}{c|}{(mIoU/\%)} & (mIoU/\%) & (mIoU/\%) & \multicolumn{1}{c|}{(mIoU/\%)} & (mIoU/\%) &(mIoU/\%) &\multicolumn{1}{c|}{(mIoU/\%)}& (mIoU/\%)&(mIoU/\%)&\multicolumn{1}{c|}{(mIoU/\%)} &(mIoU/\%) &(mIoU/\%) & (mIoU/\%)\\
      \midrule
      Baseline                                                                         & 42.06 & 40.42   & \multicolumn{1}{c|}{28.66}  & 49.66 & 70.41   & \multicolumn{1}{c|}{60.53} & 45.37     & 79.16      & 85.11    & 49.35 & 30.60   & \multicolumn{1}{c|}{26.11} & 47.56    & 22.75      & 17.23    \\ Learnable~\cite{mao2022towards}                                                                        & 42.19 & 33.42   & \multicolumn{1}{c|}{21.92} & 45.90 & 75.50   & \multicolumn{1}{c|}{66.85} & 45.41     & 80.29      & 86.98    & 49.33 & 29.38   & \multicolumn{1}{c|}{26.32} & 45.50    & 15.56      & 12.51    \\
      Temperature~\cite{gu2022are}                                                                      & 41.94 & 32.02   & \multicolumn{1}{c|}{23.12}  & 44.83 & 57.01   & \multicolumn{1}{c|}{48.41} & 44.14     & 75.09      & 81.76    & 45.86 & 22.48   & \multicolumn{1}{c|}{19.52} & 47.38    & 18.04      & 14.43    \\ \midrule
      RAM (ours)                                                                        & 42.04 & \textbf{14.59}   & \multicolumn{1}{c|}{\textbf{7.09}}  & 49.11 &  \textbf{53.76}    & \multicolumn{1}{c|}{\textbf{45.76}}      & 44.65     & \textbf{51.70} &  \textbf{48.27}        & 48.64 & \textbf{18.30}   & \multicolumn{1}{c|}{\textbf{14.92}} & 46.81    & \textbf{14.31}      & \textbf{9.88}     \\ 
      \bottomrule
      \end{tabular}
  }
  \label{tab:comparison}
\end{table*}

  We compare the effect of our proposed RAM method and other existing methods on defending patch-based attacks with various networks. As shown in~\cref{tab:comparison}, the robustness of our proposed RAM method is consistently better than existing methods.

  It is worth noting that, compared to existing methods, our proposed RAM can significantly improve the robustness of semantic segmentation models with both the local attention mechanism and global attention mechanism, which well illustrates the generalization of our proposed method. In addition, although the attention mechanism is widely used in ViT-based models in recent years, when combined with attention-based methods with CNN as backbones (\eg, Nonlocal~\cite{wang2018nonlocal}), our method can still obviously improve model robustness, which further demonstrates the good generalization of our proposed method.

  From~\cref{tab:comparison}, we can also find that although the clean mIoU of the model may be slightly reduced after combined with RAM.
  Our proposed MAS module can reduce the impact of dirty patches on the features of other locations by compressing the value range in the attention map. However, when dealing with clean images, where no dirty patches are present, the compression of the value range of the attention map within MAS can diminish the distinction between the features of different patches, which may negatively affect the clean performance of model. Yet, our method significantly enhances the model's robustness to patch-based attacks. Considering the substantial improvement in robustness, the decrease in clean accuracy is considered negligible.
  
\begin{table*}[!h]
  \caption{The robustness of different semantic segmentation models against two patch-based targeted attack settings. \textbf{clean} represents the pixel accuracy (pAcc) of the semantic segmentation results on the validation set, and \textbf{Permute} and \textbf{Strip} represent the pAcc between the segmentation results of the adversarial images and the ground truth labels of the clean images. The higher pAcc regards to \textbf{clean} means the higher segmentation accuracy of the model for clean images, and the higher pAcc regards to \textbf{Permute} and \textbf{Strip} means the better robustness of the model. The attack is conducted by DAG~\cite{xie2017adversarial}.}
  \resizebox{\linewidth}{!}{
    \begin{tabular}{c|cccccc|cccccc}
    \toprule
    Model           & \multicolumn{6}{c|}{Global Attention}                                                                          & \multicolumn{6}{c}{Local Attention}                                                                            \\ \cmidrule{2-13} 
    (Head/Backbone) & \multicolumn{3}{c|}{Nonlocal~\cite{wang2018nonlocal}/R50~\cite{he2016deep}}                                & \multicolumn{3}{c|}{SegFormer~\cite{xie2021segformer}/MiT~\cite{xie2021segformer}}          & \multicolumn{3}{c|}{SeMask~\cite{jain2021semask}/Swin~\cite{liu2021swin}}                                 & \multicolumn{3}{c}{UPerNet~\cite{xiao2018unified}/Swin~\cite{liu2021swin}}            \\ \midrule
    & clean$\uparrow$     & Permute$\uparrow$        & \multicolumn{1}{c|}{Strip$\uparrow$}          & clean$\uparrow$     & Permute$\uparrow$        & Strip$\uparrow$          & clean$\uparrow$     & Permute$\uparrow$        & \multicolumn{1}{c|}{Strip$\uparrow$}          & clean$\uparrow$     & Permute$\uparrow$        & Strip$\uparrow$          \\
    & (pAcc/\%) & (pAcc/\%)      & \multicolumn{1}{c|}{(pAcc/\%)}      & (pAcc/\%) & (pAcc/\%)      & (pAcc/\%)      & (pAcc/\%) & (pAcc/\%)      & \multicolumn{1}{c|}{(pAcc/\%)}      & (pAcc/\%) & (pAcc/\%)      & (pAcc/\%)      \\ \midrule
    Baseline        & 80.22     & 9.10           & \multicolumn{1}{c|}{9.43}           & 82.10     & 7.57           & 3.63           & 82.96     & 34.53          & \multicolumn{1}{c|}{36.42}          & 82.33     & 41.60          & 44.61          \\
    Learnable~\cite{mao2022towards}       & 80.39     & 10.18          & \multicolumn{1}{c|}{11.50}          & 82.71     & 3.82           & 3.87           & 83.13     & 35.15          & \multicolumn{1}{c|}{37.06}          & 81.57     & 46.03          & 47.92          \\
    Temperature~\cite{gu2022are}     & 80.05     & 21.86          & \multicolumn{1}{c|}{27.45}          & 82.25     & 3.22           & 2.97           & 81.73     & 41.88          & \multicolumn{1}{c|}{43.73}          & 80.77     & 46.24          & 48.59          \\ \midrule
    RAM (ours)      & 80.32     & \textbf{25.57} & \multicolumn{1}{c|}{\textbf{31.47}} & 82.22     & \textbf{13.94} & \textbf{15.39} & 82.72     & \textbf{45.12} & \multicolumn{1}{c|}{\textbf{46.91}} & 82.27     & \textbf{49.48} & \textbf{51.97} \\ \bottomrule
    \end{tabular}
  }
  \label{tab:pacc}
\end{table*}

  In order to fully demonstrate the effectiveness of our method, we also use the metric of pixel accuracy (pAcc) to evaluate the robustness of different models in~\cref{tab:pacc}. We use the pAcc to evaluate the semantic segmentation model for clean images, which denotes as clean in the table. Under the two attack settings of Permute and Strip, we also use the pAcc between the segmentation results of the adversarial images and the ground truth labels of the clean images to evaluate the robustness of models. Here, the higher pAcc means that the model is less affected by the adversarial patch, which means better model robustness.

  From ~\cref{tab:pacc}, we can clearly see that under the metric of pAcc, among all the models, our proposed RAM method can also consistently improve the robustness of semantic segmentation models against patch-based attacks, especially for the models with the global attention mechanism.

\subsubsection{Untargeted Attack Setting}
  We also conduct experiments to compare the robustness of different models under the untargeted attack setting. We utilize adversarial patches of size $100\times 100$, positioned at the bottom right corner of the images. Our experiments are conducted on the Cityscapes dataset~\cite{cordts2016the} using the DAG attack method~\cite{xie2017adversarial} to generate adversarial examples. We use the metric of the mIoU between the segmentation results of the adversarial images and the ground truth labels. The higher mIoU means the better robustness of the model. The results are shown in~\cref{tab:ground_truth}. From the results presented in the table, it's observed that our proposed method achieves an improvement of over 10\% in mIoU compared to the baseline across different models. Compared to existing defense methods, our approach also demonstrates a significant improvement in robustness, verifying the effectiveness of our method again.

\begin{table*}[]
  \centering
  \caption{The robustness of different semantic segmentation models against patch-based \textbf{untargeted attack settings}. \textbf{clean} represents the mIoU of the semantic segmentation results on the validation set, and \textbf{attack} represent the mIoU between the segmentation results of the adversarial images and the ground truth labels of the clean images. The higher mIoU regards to \textbf{clean} means the higher segmentation accuracy of the model for clean images, and the higher mIoU regards to \textbf{attack} means the better robustness of the model. The attack is conducted by DAG~\cite{xie2017adversarial}.}
  \resizebox{\linewidth}{!}{
    \begin{tabular}{c|cc|cc|cc|cc}
      \toprule
      Model & \multicolumn{4}{c|}{Global Attention} & \multicolumn{4}{c}{Local Attention} \\
      \cmidrule{2-9}(Head/Backbone) & \multicolumn{2}{c|}{Nonlocal~\cite{wang2018nonlocal}/R50~\cite{he2016deep}} & \multicolumn{2}{c|}{SegFormer~\cite{xie2021segformer}/MiT~\cite{xie2021segformer}} & \multicolumn{2}{c|}{SeMask~\cite{jain2021semask}/Swin~\cite{liu2021swin}} & \multicolumn{2}{c}{UPerNet~\cite{xiao2018unified}/Swin~\cite{liu2021swin}} \\
      \midrule
            & clean$\uparrow$ & attack$\uparrow$ & clean$\uparrow$ & attack$\uparrow$ & clean$\uparrow$ & attack$\uparrow$ & clean$\uparrow$ & attack$\uparrow$ \\
            & (mIoU/\%) & (mIoU/\%) & (mIoU/\%) & (mIoU/\%) & (mIoU/\%) & (mIoU/\%) & (mIoU/\%) & (mIoU/\%) \\
      \midrule
      Baseline & 76.12  & 25.49  & 78.56  & 23.62  & 79.10  & 37.81  & 78.91  & 35.62  \\
      Learnable~\cite{mao2022towards} & 75.37  & 24.63  & 77.82  & 25.47  & 77.32  & 38.74  & 78.01  & 36.32  \\
      Temperature~\cite{gu2022are} & 75.42  & 27.52  & 77.93  & 24.92  & 77.19  & 39.93  & 77.42  & 39.17  \\
      \midrule
      RAM (ours) & 75.91  & \textbf{35.56} & 78.03  & \textbf{32.67} & 78.09  & \textbf{50.97} & 77.78  & \textbf{48.31} \\
      \bottomrule
    \end{tabular}
  }
  \label{tab:ground_truth}
\end{table*}

\subsection{Robustness against Different Attack Methods}

\begin{table*}[]
  \centering
  \caption{The robustness of semantic segmentation models against different patch-based attack methods.
  The results in the table are the mIoU between the segmentation results of the adversarial images and the target labels determined by the adversary.
  }
  \resizebox{\linewidth}{!}{
  \begin{tabular}{c|cc|cc|cc|cc|cc}
  \toprule
  \multirow{3}{*}{} & \multicolumn{2}{c|}{PGD~\cite{madry2018towards}} & \multicolumn{2}{c|}{DAG~\cite{xie2017adversarial}} & \multicolumn{2}{c|}{IPatch~\cite{mirsky2021ipatch}} & \multicolumn{2}{c|}{SSAP~\cite{nesti2022evaluating}} & \multicolumn{2}{c}{Patch-Fool~\cite{fu2022patch}} \\ \cmidrule{2-11} 
  & Permute$\downarrow$      & Strip$\downarrow$     & Permute$\downarrow$      & Strip$\downarrow$     & Permute$\downarrow$     & Strip$\downarrow$     & Permute$\downarrow$      & Strip$\downarrow$  & Permute$\downarrow$      & Strip$\downarrow$ \\
  & (mIoU/\%) & (mIoU/\%) & (mIoU/\%) & (mIoU/\%) & (mIoU/\%) & (mIoU/\%) & (mIoU/\%) & (mIoU/\%) & (mIoU/\%) & (mIoU/\%) \\
  \midrule
  SegFormer/MiT~\cite{xie2021segformer} & 54.73 & 47.80 & 49.32 & 40.01 & 53.21 & 46.55 & 39.96 & 28.45 & 57.21 & 46.55  \\
  RAM (ours) & \textbf{37.53} & \textbf{26.13} & \textbf{33.72} & \textbf{22.69} & \textbf{36.10} & \textbf{24.51} & \textbf{26.11} & \textbf{14.68} & \textbf{35.02} & \textbf{24.67} \\ \midrule
  UPerNet~\cite{xiao2018unified}/DeiT~\cite{touvron2021training} & 81.64 & 88.26 & 79.16 & 85.11 & 79.45 & 87.47 & 44.15 & 37.80 & 82.41 & 87.29  \\
  RAM (ours) & \textbf{57.12} & \textbf{56.34} & \textbf{51.70} & \textbf{48.27} & \textbf{54.65} & \textbf{53.77} & \textbf{20.20} & \textbf{13.32} & \textbf{57.25} & \textbf{55.68}  \\ \midrule
  SeMask~\cite{jain2021semask}/Swin~\cite{liu2021swin} & 30.58 & 30.47 & 30.60 & 26.11 & 29.25 & 28.97 & 22.74 & 20.53 & 31.73 & 30.48 \\
  RAM (ours) & \textbf{19.32} & \textbf{16.92} & \textbf{18.30} & \textbf{14.92} & \textbf{18.51} & \textbf{15.86} & \textbf{13.87} & \textbf{11.16} & \textbf{19.91} & \textbf{17.36}  \\
  \bottomrule
  \multirow{3}{*}{} & \multicolumn{2}{c|}{Attention-Fool~\cite{lovisotto2022give}} & \multicolumn{2}{c|}{AutoAttack~\cite{croce2020reliable}} & \multicolumn{2}{c|}{EOT~\cite{athalye2018synthesizing}} & \multicolumn{2}{c|}{MaxVarDAG} & \multicolumn{2}{c}{MaxAttnDAG}\\ \cmidrule{2-11} 
  & Permute$\downarrow$     & Strip$\downarrow$     & Permute$\downarrow$      & Strip$\downarrow$  & Permute$\downarrow$      & Strip$\downarrow$ & Permute$\downarrow$      & Strip$\downarrow$ & Permute$\downarrow$      & Strip$\downarrow$    \\
  & (mIoU/\%) & (mIoU/\%) & (mIoU/\%) & (mIoU/\%) & (mIoU/\%) & (mIoU/\%) & (mIoU/\%) & (mIoU/\%) & (mIoU/\%) & (mIoU/\%) \\
  \midrule
  SegFormer/MiT~\cite{xie2021segformer} & 55.19 & 45.91 & 58.43 & 52.69 & 49.08 & 45.19 & 45.26 & 40.92 & 47.33 & 37.89 \\
  RAM (ours) & \textbf{34.12} & \textbf{23.95} & \textbf{39.14} & \textbf{29.97} &  \textbf{31.28} & \textbf{20.64} & \textbf{30.84} & \textbf{21.26} & \textbf{31.15} & \textbf{22.21}  \\ \midrule
  UPerNet~\cite{xiao2018unified}/DeiT~\cite{touvron2021training} & 80.93 & 88.65 & 84.48 & 91.31 & 72.09 & 75.01 & 67.81 & 71.64 & 69.16 & 73.84 \\
  RAM (ours) & \textbf{57.03} & \textbf{57.33} & \textbf{60.37} & \textbf{62.62} & \textbf{49.42} & \textbf{51.73} & \textbf{37.12} & \textbf{40.75} & \textbf{39.10} & \textbf{41.78} \\ \midrule
  SeMask~\cite{jain2021semask}/Swin~\cite{liu2021swin} & 29.62 & 28.08 & 33.94 & 32.91 & 26.13 & 23.78 & 27.18 & 24.56 & 29.41 & 27.23\\
  RAM (ours) & \textbf{17.20} & \textbf{15.16} & \textbf{21.07} & \textbf{20.79} & \textbf{12.98} & \textbf{11.97} & \textbf{18.61} & \textbf{17.38} & \textbf{14.07} & \textbf{13.14} \\
  \bottomrule
  \end{tabular}
  }
  \label{tab:all_attack}
\end{table*}

  To further investigate the effectiveness of our RAM, we introduce various attack methods for evaluation to compare the robustness of the semantic segmentation models between the baseline models and our proposed RAM against patch-based attacks. In order to comprehensively evaluate the defense effect of our proposed RAM, we first use the widely used PGD~\cite{madry2018towards} method, together with several strong attack methods designed for semantic segmentation models (\ie, DAG~\cite{xie2017adversarial}, IPatch~\cite{mirsky2021ipatch}, SSAP~\cite{nesti2022evaluating}) to evaluate the robustness of models. In addition, we further adopt recent proposed attention-based attack methods (\eg, Patch-Fool~\cite{fu2022patch} and Attention-Fool~\cite{lovisotto2022give}), together with sevwral strong attack methods which often used as the benchmark (\ie, AutoAttack~\cite{croce2020reliable} and EOT~\cite{athalye2018synthesizing}), to further evaluate the robustness of segmentation models.
  Further, in order to more comprehensively evaluate the effectiveness of our defense method, we use several adaptive attack methods to conduct the evaluation. Specifically, based on the DAG method, we propose two adaptive attack methods, MaxVarDAG and MaxAttnDAG to elaborately attack the attention matrix. In the~\cref{sec:att_method}, we provide a detailed introduction to all the attack methods used.
  
  As shown in~\cref{tab:all_attack}, our proposed RAM achieves consistently better robustness under all attack methods with an average robustness improvement of nearly 20\% compared to the baseline, which demonstrates the superiority of our proposed RAM.

\subsection{Comparison with Adversarial Training}
  We compare our proposed defense method with the method of adversarial training under two targeted attack settings. We use the method of PGD~\cite{madry2018towards} with 10 attack steps to conduct the adversarial training. It should be noted that, although adversarial training is a widely used and effective defense method, the training time and computing resources required by adversarial training far exceed our method of refining the attention mechanism. From~\cref{tab:adversarial_train}, it can be seen from the table that under almost all attack methods, our method achieves better robustness than the adversarial training method, which further demonstrates the effectiveness of our proposed defense method.

\begin{table*}[b]
  \centering
  \caption{Comparison with adversarial training.}
  \resizebox{\linewidth}{!}{
    \begin{tabular}{c|cc|cc|cc|cc|cc}
    \toprule
    \multirow{3}{*}{} & \multicolumn{2}{c|}{PGD~\cite{madry2018towards}} & \multicolumn{2}{c|}{DAG~\cite{xie2017adversarial}} & \multicolumn{2}{c|}{IPatch~\cite{mirsky2021ipatch}} & \multicolumn{2}{c|}{SSAP~\cite{nesti2022evaluating}} & \multicolumn{2}{c}{Patch-Fool~\cite{fu2022patch}} \\ \cmidrule{2-11} 
    & Permute$\downarrow$      & Strip$\downarrow$     & Permute$\downarrow$      & Strip$\downarrow$     & Permute$\downarrow$     & Strip$\downarrow$     & Permute$\downarrow$      & Strip$\downarrow$  & Permute$\downarrow$      & Strip$\downarrow$ \\
    & (mIoU/\%) & (mIoU/\%) & (mIoU/\%) & (mIoU/\%) & (mIoU/\%) & (mIoU/\%) & (mIoU/\%) & (mIoU/\%) & (mIoU/\%) & (mIoU/\%) \\
    \midrule
    SegFormer/MiT~\cite{xie2021segformer} & 54.73 & 47.80 & 49.32 & 40.01 & 53.21 & 46.55 & 39.96 & 28.45 & 57.21 & 46.55  \\
    Adversarial Training & \textbf{37.19} & 27.10 & 39.72 & 29.60 & 42.99 & 35.27 & 27.65 & 14.97 & 43.67 & 31.24 \\
    RAM (ours) & 37.53 & \textbf{26.13} & \textbf{33.72} & \textbf{22.69} & \textbf{36.10} & \textbf{24.51} & \textbf{26.11} & \textbf{14.68} & \textbf{35.02} & \textbf{24.67} \\
    \bottomrule
    \multirow{3}{*}{} & \multicolumn{2}{c|}{Attention-Fool~\cite{lovisotto2022give}} & \multicolumn{2}{c|}{AutoAttack~\cite{croce2020reliable}} & \multicolumn{2}{c|}{EOT~\cite{athalye2018synthesizing}} & \multicolumn{2}{c|}{MaxVarDAG} & \multicolumn{2}{c}{MaxAttnDAG}\\ \cmidrule{2-11} 
    & Permute$\downarrow$     & Strip$\downarrow$     & Permute$\downarrow$      & Strip$\downarrow$  & Permute$\downarrow$      & Strip$\downarrow$ & Permute$\downarrow$      & Strip$\downarrow$ & Permute$\downarrow$      & Strip$\downarrow$    \\
    & (mIoU/\%) & (mIoU/\%) & (mIoU/\%) & (mIoU/\%) & (mIoU/\%) & (mIoU/\%) & (mIoU/\%) & (mIoU/\%) & (mIoU/\%) & (mIoU/\%) \\
    \midrule
    SegFormer/MiT~\cite{xie2021segformer} & 55.19 & 45.91 & 58.43 & 52.69 & 49.08 & 45.19 & 45.26 & 40.92 & 47.33 & 37.89 \\
    Adversarial Training & 41.09 & 33.43 & 45.87 & 37.18 & 40.27 & 36.19 & 37.12 & 32.76 & 38.03 & 31.14 \\
    RAM (ours) & \textbf{34.12} & \textbf{23.95} & \textbf{39.14} & \textbf{29.97} &  \textbf{31.28} & \textbf{20.64} & \textbf{30.84} & \textbf{21.26} & \textbf{31.15} & \textbf{22.21}  \\ 
    \bottomrule
    \end{tabular}
  }
  \label{tab:adversarial_train}
\end{table*}

We also compare our RAM with advanced adversarial training method~\cite{xu2021dynamic} under the untargeted attack setting in~\cref{tab:advanced_adversarial_train}.
From the table we can find that the robustness of the DDC-AT method~\cite{xu2021dynamic} is similar to our proposed RAM, but it is important to note that DDC-AT require approximately three times the training time compared to our method. Additionally, our proposed method can be seamlessly tailored into adversarial training approaches to further boost the robustness of semantic segmentation models against patch-based attacks. As seen in the table, combining our RAM with DDC-AT further improves the robustness of semantic segmentation models to patch-based attacks.

\begin{table*}[b]
  \centering
  \caption{Integration with advanced adversarial training method.}
  \resizebox{\linewidth}{!}{
    \begin{tabular}{c|cc|cc|cc|cc}
      \toprule
      Model & \multicolumn{4}{c|}{Global Attention} & \multicolumn{4}{c}{Local Attention} \\
      \cmidrule{2-9}(Head/Backbone) & \multicolumn{2}{c|}{Nonlocal~\cite{wang2018nonlocal}/R50~\cite{he2016deep}} & \multicolumn{2}{c|}{SegFormer~\cite{xie2021segformer}/MiT~\cite{xie2021segformer}} & \multicolumn{2}{c|}{SeMask~\cite{jain2021semask}/Swin~\cite{liu2021swin}} & \multicolumn{2}{c}{UPerNet~\cite{xiao2018unified}/Swin~\cite{liu2021swin}} \\
      \midrule
            & clean$\uparrow$ & attack$\uparrow$ & clean$\uparrow$ & attack$\uparrow$ & clean$\uparrow$ & attack$\uparrow$ & clean$\uparrow$ & attack$\uparrow$ \\
            & (mIoU/\%) & (mIoU/\%) & (mIoU/\%) & (mIoU/\%) & (mIoU/\%) & (mIoU/\%) & (mIoU/\%) & (mIoU/\%) \\
      \midrule
      Baseline & 76.12  & 25.49  & 78.56  & 23.62  & 79.10  & 37.81  & 78.91  & 35.62  \\
      DDC-AT~\cite{xu2021dynamic} & 75.21  & 34.78  & 76.86  & 32.88  & 77.10  & 49.69  & 76.28  & 48.16  \\
      \midrule
      RAM (ours) & 75.91  & 35.56  & 78.03  & 32.67  & 78.09  & 50.97  & 77.78  & 48.31  \\
      DDC-AT~\cite{xu2021dynamic} + RAM (ours) & 75.37  & \textbf{37.81} & 77.74  & \textbf{33.52} & 77.65  & \textbf{52.43} & 77.46  & \textbf{50.97} \\
      \bottomrule
    \end{tabular}%
  }
  \label{tab:advanced_adversarial_train}
\end{table*}

\subsection{Results on More Datasets}
  To further demonstrate the excellent effectiveness of our proposed RAM, we also conduct experiments on the VOC2012~\cite{everingham2015the} and Cityscapes~\cite{cordts2016the} datasets. It can be clearly seen from~\cref{tab:more_dataset} that our proposed RAM also significantly improves the robustness of the model against patch-based attacks on the VOC2012 and Cityscapes datasets. Especially for the Cityscapes dataset, the introduction of RAM almost halves the attack success rate under targeted attacks, which fully demonstrates the effectiveness of the RAM in improving the robustness of the semantic segmentation model.

\begin{table}[!t]
   \centering
   \caption{The robustness of semantic segmentation models on VOC2012 and Cityscapes.}
       \begin{tabular}{c|c|cc}
       \toprule
       \multirow{2}*{Dataset} & \multirow{2}*{Model}& Permute$\downarrow$  & Strip$\downarrow$      \\
       & & (mIoU/\%) & (mIoU/\%) \\
       \midrule
       \multirow{2}*{VOC2012~\cite{everingham2015the}}   & Nonlocal~\cite{wang2018nonlocal}/R50~\cite{he2016deep} & 82.02 & 23.12   \\
       &  RAM (ours) & \textbf{56.81} & \textbf{9.56}  \\ 
       \midrule
       \multirow{2}*{Cityscapes~\cite{cordts2016the}}   & SegFormer/MiT~\cite{xie2021segformer} & 87.06 & 90.88   \\
       &  RAM (ours) & \textbf{46.58} & \textbf{42.39}  \\ \bottomrule
       \end{tabular}

   \label{tab:more_dataset}
\end{table}

\subsection{Visualization}
\begin{figure}[!t]
  \centering
  \includegraphics[width=\columnwidth]{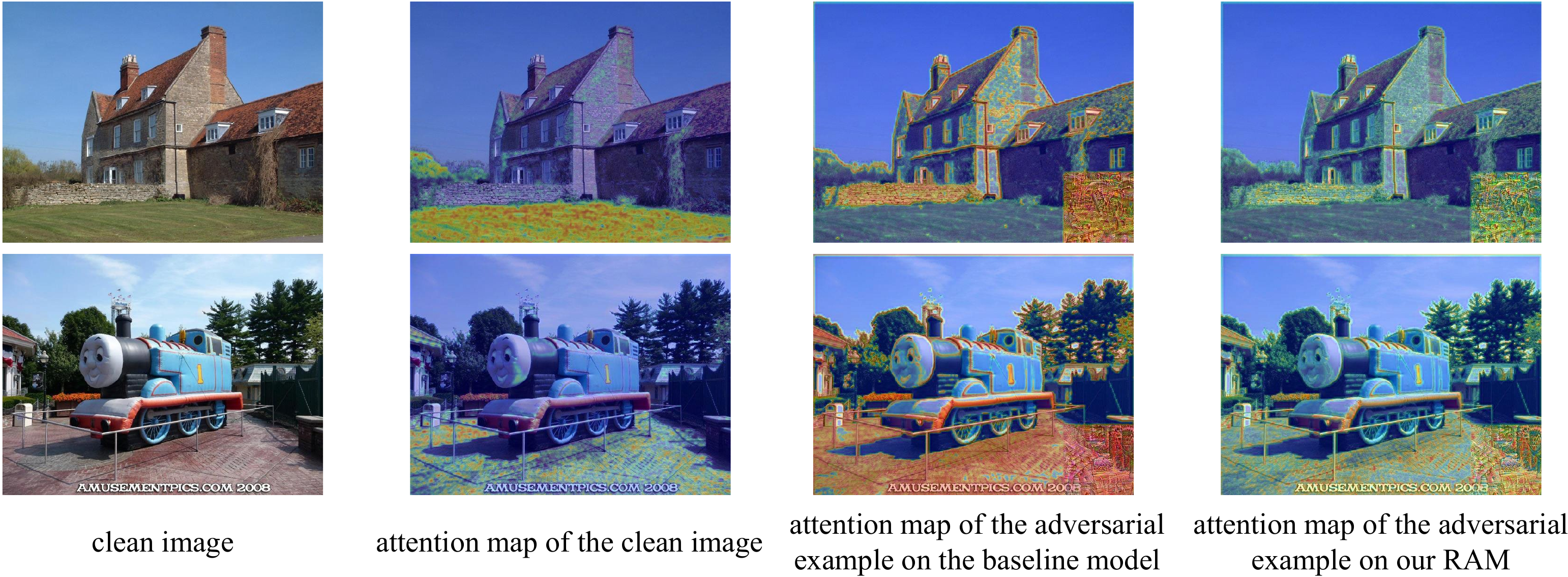}
  \caption{Visualization of attention maps, where the first column shows the clean image, the second column shows the attention map of the clean image, the third and fourth columns show the attention maps of the feature at the adversarial patch location to all locations in the image, generated by the baseline method and our proposed RAM method, respectively. Redder colors indicate greater influence.}
  \label{fig:visual_attention}
\end{figure}

  We compare the visualizations of attention maps when facing adversarial examples between our proposed RAM method and baseline methods, as shown in~\cref{fig:visual_attention}. We conduct experiments on the ADE20K dataset. We compare the attention maps of the feature at the adversarial patch location to all locations in the image, generated using different methods. In the figure, the first column shows the clean image, the second column shows the attention map of the clean image, the third and fourth columns show the attention maps of the feature at the adversarial patch location to all locations in the image, generated by the baseline method and our proposed RAM method, respectively. After converting the attention map into a heatmap, we overlay it on the corresponding adversarial example with an alpha of 0.3 for visualization purposes. Redder colors in the foreground indicate greater influence.
  From the figure, it is evident that in the clean image, local patches that are semantically relevant to specific regions exhibit greater attention, whereas regions that are semantically irrelevant display lesser attention. In contrast, the influence of the adversarial patch spreads to all areas, regardless of their semantic relevance to the region’s content. Additionally, we can observe that our proposed RAD module, by suppressing the value range of the attention map, significantly reduces the impact of adversarial patch features on the features of other clean areas compared to baseline methods. Therefore, it can enhance the robustness of semantic segmentation models to patch-based attacks.

\begin{figure}[!t]
  \centering
  \includegraphics[width=\columnwidth]{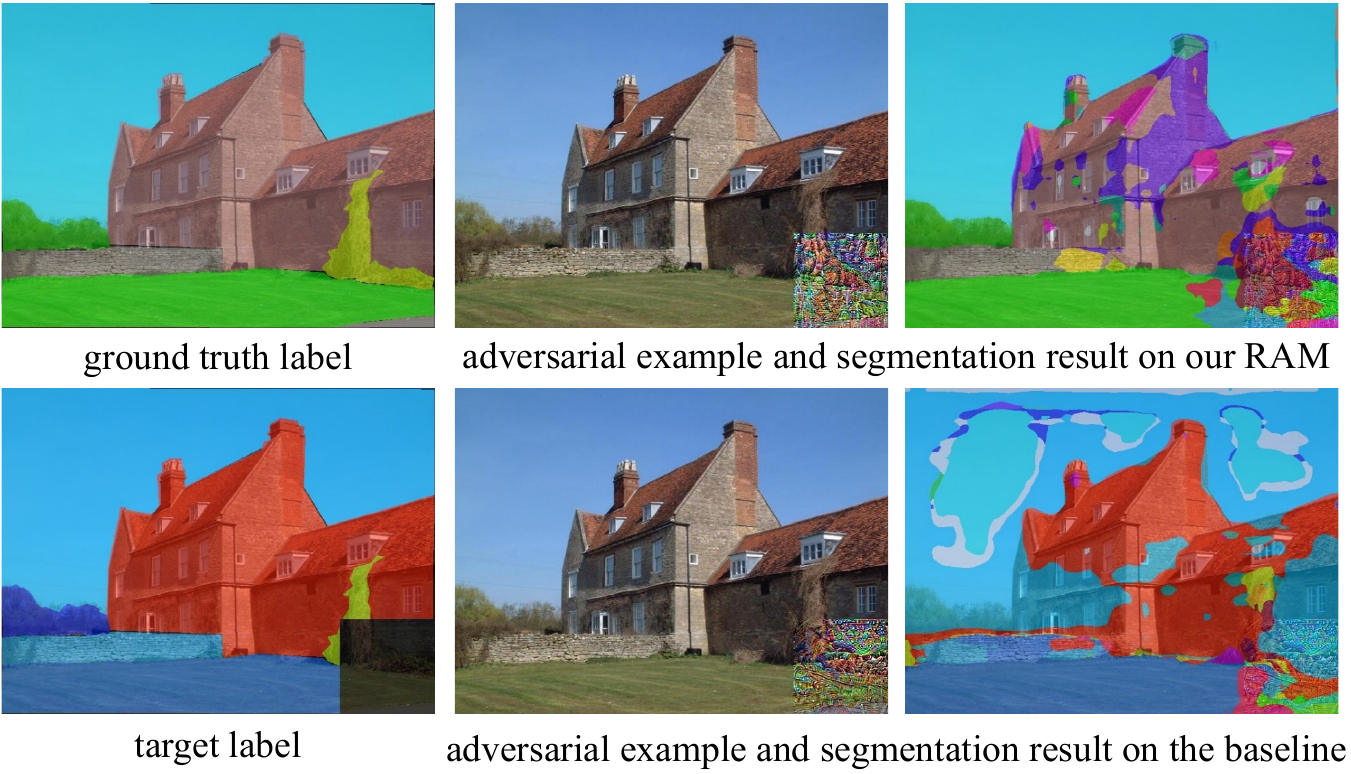}
  \caption{Visualization of the segmentation results of Nonlocal~\cite{wang2018nonlocal}/R50~\cite{he2016deep} model with our proposed RAM and the baseline model under the attack setting of Permute.}
  \label{fig:visual_nonlocal}
\end{figure}

  We also visualize the segmentation results of Nonlocal~\cite{wang2018nonlocal}/R50~\cite{he2016deep} model with our proposed RAM and the baseline model under the attack setting of Permute. As shown in~\cref{fig:visual_nonlocal}, most of the segmentation results towards the adversarial examples of our proposed RAM model still maintain the ground truth label.
  In contrast, the adversarial example almost completely modifies the output of the baseline model, and the segmentation results are very close to the target labels specified by the adversary. It demonstrates that the baseline model is quite vulnerable to patch-based attacks, while our RAM method significantly improves the robustness of semantic segmentation models against attacks. More visualization results are shown in~\cref{sec:visual}.

\section{Conclusion}
\label{sec:conclusion}
  In this work, we study the robustness of semantic segmentation against patch-based attacks. Through the analysis of the effective receptive field, we find that the model with a wider receptive field is more vulnerable to patch-based attacks, and the self-attention module is the key to the robustness of the model. To improve the robustness of the attention-based semantic segmentation model, we propose a Robust Attention Mechanism (RAM), which introduces two novel modules called Max Attention Suppression and Random Attention Dropout, both of which aim to constrain the influence of a single adversarial patch on the semantic segmentation results of other regions.
  Extensive experiments demonstrate the effectiveness of our proposed methods to improve the robustness against various patch-based attack methods under different attack settings.
  In the future, we will further explore methods that can effectively improve the robustness of models against patch-based attacks in other vision tasks.

\backmatter








\bibliography{sn-bibliography}

\newpage
\begin{appendices}

\section{The Details of Attack Methods}
\label{sec:att_method}
  In this section, we first introduce the general framework of attack methods, and then describe ten attack methods (\ie, PGD~\cite{madry2018towards}, DAG~\cite{xie2017adversarial}, IPatch~\cite{mirsky2021ipatch}, SSAP~\cite{nesti2022evaluating}, Patch-Fool~\cite{fu2022patch}, Attention-Fool~\cite{lovisotto2022give}, AutoAttack~\cite{croce2020reliable}, EOT~\cite{athalye2018synthesizing}, MaxVarDAG and MaxAttnDAG) in detail.

  In order to comprehensively evaluate the defense effect of our proposed RAM, we first use the widely used PGD~\cite{madry2018towards} method, together with three strong attack methods designed for semantic segmentation models (\ie, DAG~\cite{xie2017adversarial}, IPatch~\cite{mirsky2021ipatch}, SSAP~\cite{nesti2022evaluating}) to evaluate the robustness of models. In addition, we further adopt recent proposed attention-based attack methods (\eg, Patch-Fool~\cite{fu2022patch} and Attention-Fool~\cite{lovisotto2022give}), together with several strong attack methods which often used as the benchmark (\ie, AutoAttack~\cite{croce2020reliable} and EOT~\cite{athalye2018synthesizing}), to further evaluate the robustness of segmentation models. Moreover, in order to more comprehensively evaluate the effectiveness of our defense method, we use several adaptive attack methods to conduct the evaluation. Specifically, based on the DAG method, we propose two adaptive attack methods, MaxVarDAG and MaxAttnDAG to elaborately attack the attention matrix.

\subsection{The General Framework of Attack Methods}
  We use the process of adversarial attack against the semantic segmentation model in DAG~\cite{xie2017adversarial} method as the general framework, and add the limitation of a patch-based mask to it. The algorithm is shown in~\cref{alg:attack}, where $T$ is the number of iterative steps in the attack, $\gamma$ is a hyperparameter to control the scale of the gradient, $\text{Clip}$ means to clip the generated adversarial examples to the image scale range, \ie, $[0,1]$. The specific loss functions used in different attack methods are introduced below.

  In the experiments, we set $T=400$ for all attack methods. For the hyperparameter of $\gamma$, we set $\gamma=0.005$ for PGD~\cite{madry2018towards}, and $\gamma=1$ for the other all attack methods.

\begin{algorithm}[!h]    
  \algnewcommand\algorithmicinput{\textbf{Input:}}
  \algnewcommand\Input{\item[\algorithmicinput]}
  \algnewcommand\algorithmicoutput{\textbf{Output:}}
  \algnewcommand\Output{\item[\algorithmicoutput]}
  \caption{The General Framework of Attack Methods}                 
  \begin{algorithmic}[1]   
    \Input the clean image $\mathbf{x}$, the patch-based mask $\mathbf{M_{mask}}$
    \Output the patch-based adversarial examples $\mathbf{x^{adv}}$         
    \State $\mathbf{x_0 = x}$
    \For {i $\in \{0, \cdots, T-1\}$ }
      \State Calculate the loss function $\mathcal{L}$ according different attack methods
      \State $\mathbf{r_i}=\frac{\partial \mathcal{L}}{\partial \mathbf{x_i}}$
      \State $\mathbf{r'_i}=\frac{\gamma}{\|\mathbf{r_i}\|_{\infty}} \mathbf{r_i}\odot \mathbf{M_{mask}}$
      \State $\mathbf{x_{i+1}}=\text{Clip}(\mathbf{x_i+r'_i})$
    \EndFor
    \State $\mathbf{x^{adv}=x_T}$
    \State \Return $\mathbf{x^{adv}}$
  \end{algorithmic}
  \label{alg:attack}
\end{algorithm}
\subsection{Projected Gradient Descent}
  Projected Gradient Descent (PGD)~\cite{madry2018towards}
  method is one of the earliest classic adversarial attack methods, and it focuses on the adversarial attack of the classification model under $L_p$ norm. In the targeted attack scenario, it generates adversarial examples by maximizing the cross-entropy loss between model predictions and target labels. In this paper, we migrate it to the task of patch-based attacks on semantic segmentation models. Assume that the predicted segmentation result of the model for the input image $\mathbf{X}\in \mathbb{R}^{H\times W\times 3}$ is $\mathbf{Y}\in[0,1]^{H\times W\times C}$, the ground truth label is $\mathbf{Y^g}\in \{0,1,\cdots, C-1\}^{H\times W}$ and the target label specified by the attacker is $\mathbf{Y^t}\in \{0,1,\cdots, C-1\}^{H\times W}$, where $H$ and $W$ are the height and width of the input image, and $C$ is the number of categories for image segmentation. The loss function of PGD method is as follows:
  \begin{equation}
    \mathcal{L}_{\text{PGD}}=\sum_{i=0}^{H-1}\sum_{j=0}^{W-1}\log \mathbf{Y}_{ij\mathbf{Y^t}_{ij}},
  \end{equation}
  where $\mathbf{Y^t}_{ij}$ represents the target label at position $(i, j)$ in the image, and $\mathbf{Y}_{ijc}$ represents the probability predicted by the model that the pixel at position $(i,j)$ in the image belongs to category $c$.

\subsection{Dense Adversary Generation}

  Dense Adversary Generation (DAG)~\cite{xie2017adversarial} method is an attack method specific to dense prediction tasks, such as semantic segmentation and object detection. The basic idea
  is to define a dense set of targets as well as a different
  set of desired labels, and optimize a loss function in order
  to produce incorrect recognition results on all the targets
  simultaneously. The loss function used in DAG is as follows:
  \begin{gather}
    \mathcal{T} = \{(i,j)|\arg\max_c \mathbf{Y}_{ijc} = \mathbf{Y^g}_{ij} \}, \\
    \mathcal{L}_{\text{DAG}}=\sum_{(i,j)\in\mathcal{T}} \mathbf{Y}_{ij\mathbf{Y^g}_{ij}} - \mathbf{Y}_{ij\mathbf{Y^t}_{ij}}.
  \end{gather}

\subsection{IPatch}
  IPatch~\cite{mirsky2021ipatch} aims to change the semantics of locations far from the patch to a specific category by placing a universal adversarial patch anywhere within an image. When we utilize the method of IPatch to conduct the patch-based attack in our work, we fix the location of the adversarial patch at the lower right corner of the image and generate the adversarial patch for each image separately. The loss function used in IPatch is as follows:
  \begin{align}
    \mathcal{L}_{\text{IPatch}}&=\mathcal{L}_{\text{KL}}(\text{onehot}(\mathbf{Y^t}), \mathbf{Y})\\
    &=\sum_{i=0}^{H-1}\sum_{j=0}^{W-1}\sum_{c=0}^{C-1}\text{onehot}(\mathbf{Y^t})_{ijc}\cdot \log\frac{\text{onehot}(\mathbf{Y^t})_{ijc}+\epsilon}{\mathbf{Y}_{ijc}},
  \end{align}
  where $\epsilon=1\mathrm{e}{-10}$.

\subsection{Scene-specific Patch Attack}
  The Scene-specific Patch Attack (SSPA)~\cite{nesti2022evaluating} method designs a new loss function, which considers the pixels that have been successfully attacked and the pixels that have not yet been successfully attacked simultaneously, and adjusts the weights of the two terms through adaptive coefficients. The attack process is formalized as follows:
  \begin{gather}
    \mathcal{T}=\{(i,j)|\arg\max_c \mathbf{Y}_{ijc}\neq \mathbf{Y^t}_{ij} \}, \\
    \eta = \frac{|\mathcal{T}|}{H\cdot W}, \\
    \mathcal{L}_{\text{SSAP}}=\eta\cdot \sum_{(i,j)\in\mathcal{T}}\log \mathbf{Y}_{ij\mathbf{Y^t}_{ij}} + (1-\eta)\cdot \sum_{(i,j)\notin \mathcal{T} }\log \mathbf{Y}_{ij\mathbf{Y^t}_{ij}}.
  \end{gather}

\subsection{Patch-Fool}
  Given the insensitivity of ViTs’ self-attention mechanism to local perturbations in the task of classification, Patch-Fool~\cite{fu2022patch} is designed to fool the self-attention mechanism by attacking its basic component (\ie, a single patch) with a series of attention-aware optimization techniques. It should be noted that the Patch-Fool method is an attack method designed for the classification model, so we have partially modified the loss function in order to be suitable for the attack on semantic segmentation models.

  The loss function used in Patch-Fool consists of two parts, one of which is the commonly used cross-entropy loss:
  \begin{equation}
    \mathcal{L}_{\text{CE}}=\sum_{i=0}^{H-1}\sum_{j=0}^{W-1}\log \mathbf{Y}_{ij\mathbf{Y^t}_{ij}}.
  \end{equation}
  The other one is the attention-aware loss:
  \begin{equation}
    \mathcal{L}_{\text{ATTN}}^{(l)}=\sum_{h,i}a_p^{(l,h,i)},
  \end{equation}
  where $\mathbf{a^{(l,h,i)}}=[a_1^{(l,h,i)}, \cdots, a_n^{(l,h,i)}]\in \mathbb{R}^n$ is the attention distribution for the $i$-th token of the $h$-th head in the $l$-th layer, $a_p^{(l,h,i)}$ denotes the attention value corresponding to the adversarial patch $p$.

  The adversarial patch is then updated based on both the cross-entropy loss and the layer-wise attention-aware loss:
  \begin{equation}
    \mathcal{L}_{\text{Patch-Fool}}=\mathcal{L}_{\text{CE}}+\alpha \sum_l \mathcal{L}_{\text{ATTN}}^{(l)} - \alpha \sum_l \beta_l \mathcal{L}_{\text{CE}},
  \end{equation}
  where 
  \begin{equation}
    \beta_l=
    \begin{cases}
    0,& \langle \nabla_\mathbf{x} \mathcal{L}_{\text{CE}}, \nabla_\mathbf{x} \mathcal{L}_{\text{ATTN}}^{(l)} \rangle >0,  \\
    \frac{\langle \nabla_\mathbf{x} \mathcal{L}_{\text{CE}}, \nabla_\mathbf{x} \mathcal{L}_{\text{ATTN}}^{(l)} \rangle}{\|\nabla_\mathbf{x} \mathcal{L}_{\text{CE}}\|^2},& \text{otherwise},
    \end{cases}
    \end{equation}
  where $l$ is the predefined self-attention layer and $\alpha$ is a weighted coeffficient.

\subsection{Attention-Fool}
  Attention-Fool~\cite{lovisotto2022give} claims that Vision Transformer (ViT) models may give a false sense of robustness due to dot-product attention, and can be easily attacked with a specific kind of attack called Attention-Fool attack. It should be noted that the Attention-Fool method is also an attack method designed for the classification model, so we have partially modified the loss function in order to be suitable for the attack on semantic segmentation models.

  We denote projected queries by $\mathbf{P_Q^{hl}=X^{hl}W_Q^{hl}}\in \mathbb{R}^{n\times d_q}$ and projected keys by $\mathbf{P_K^{hl}=X^{hl}W_K^{hl}}\in \mathbb{R}^{n\times d_k}$, where $\mathbf{X^{hl}}$ is the feature of images fed to the $h$-th attention head of $l$-th layer, $\mathbf{W_Q^{hl}}$ and $\mathbf{W_K^{hl}}$ are the learned projection matrices, respectively. Then $\mathbf{B^{hl}}=\frac{\mathbf{P_Q^{hl}}(\mathbf{P_K^{hl}})^\top}{\sqrt{d_k}}\in\mathbb{R}^{n\times n}$ quantifies the dot-product similarity between each pair of key and query in the attention head $h$ and layer $l$.

  $\mathcal{L}_{kq}$ is proposed to attack all attention layers and heads simultaneously, which is formalized as follows:
  \begin{gather}
    \mathcal{L}_{kq}^{hl}=\frac{1}{n}\sum_j \mathbf{B_{jp}^{hl}}, \\
    \mathcal{L}_{kq}=\log \sum_l \exp(\log \sum_h \exp(\mathcal{L}_{kq}^{hl})),
  \end{gather}
  where $\mathbf{B_{jp}^{hl}}$ denotes the corresponding feature of adversarial patch $p$.

  The loss function used in Attention-Fool attack is the combination of the cross-entropy loss and $\mathcal{L}_{kq}$:
  \begin{equation}
    \mathcal{L}_{\text{Attention-Fool}} = \mathcal{L}_{\text{CE}}+\alpha \mathcal{L}_{kq},
  \end{equation}
  where $\alpha$ is a weighted coefficient.

\subsection{AutoAttack}

  AutoAttack~\cite{croce2020reliable} is a strong adaptive attack method that has been widely used in recent years to evaluate the robustness of the model. It is a kind of parameter-free ensemble attack, which consists of four attack methods, \ie, APGD$_{\text{CE}}$, APGD$_{\text{DLR}}$, FAB~\cite{croce2020minimally} and Square Attack~\cite{andriushchenko2020square}. It should be noted that AutoAttack is also an attack method designed for the classification model, so we have partially modified the loss function in order to be suitable for the attack on semantic segmentation models.

\subsection{EOT}
  Expectation Over Transformation (EOT)~\cite{athalye2018synthesizing} method is first proposed by Athalye \etal, which calculates the expected value of the loss function over the distribution of possible transformations (\eg, random rotation, translation, addition of noise), and then uses this expected value to guide the generation of adversarial examples. EOT method allows the attacker to craft adversarial examples that are effective even when the defense mechanism introduces randomness into the input image.

\subsection{MaxVarDAG}
  In order to further explore the defense effect of our method under adaptive attack, we designed two additional adaptive attack methods (\ie, MaxVarDAG and MaxAttnDAG) based on DAG, that is, to elaborately attack the design of our attention mechanism, so as to more comprehensively evaluate the effectiveness of our defensive approach. Specifically, MaxVarDAG extends the DAG method by introducing an additional regularization term that maximizes the variance of the attention matrix in order to amplify the influence of the dirty patch. The loss function of MaxVarDAG is formalized as follows:
  \begin{equation}
    \mathcal{L}_{\text{MaxVarDAG}} = \mathcal{L}_{\text{DAG}}+\alpha \cdot \text{Var} (\mathbf{M}),
  \end{equation}
  where $\mathbf{M}$ is the normalized attention matrix in \cref{equ:m}, $\text{Var}$ represents the funciton of variance, $\alpha$ is a weighted coefficient.

\subsection{MaxAttnDAG}
  MaxAttnDAG also extends DAG by directly adding a regularization term that maximizes the attention value corresponding to the dirty patch position. Through this design, the inference of the dirty patch to the result of image semantic segmentation can be maximized, so as to conduct the adaptive attack based on our proposed defense method. The loss function of MaxVarDAG is formalized as follows:
  \begin{equation}
    \mathcal{L}_{\text{MaxAttnDAG}} = \mathcal{L}_{\text{DAG}}+\alpha \cdot \text{resize}(\sum_i \mathbf{M_{i\cdot}}) \odot \mathbf{M_{mask}},
  \end{equation}
  where $\mathbf{M} \in \mathbb{R}^{N\times N}$ is the normalized attention matrix in \cref{equ:m}, $\text{resize}$ refers to the process of resizing the attention value vector of each position $\sum_i \mathbf{M_{i\cdot}}$ to a new size of $H\times W$, $\mathbf{M_{mask}} \in \{0,1\}^{H\times W}$ is the mask of dirty patch, $\alpha$ is a weighted coefficient.

\begin{figure*}[!thbp]
  \centering
  \begin{subfigure}[t]{\textwidth}
    \centering
    \includegraphics[width=\textwidth]{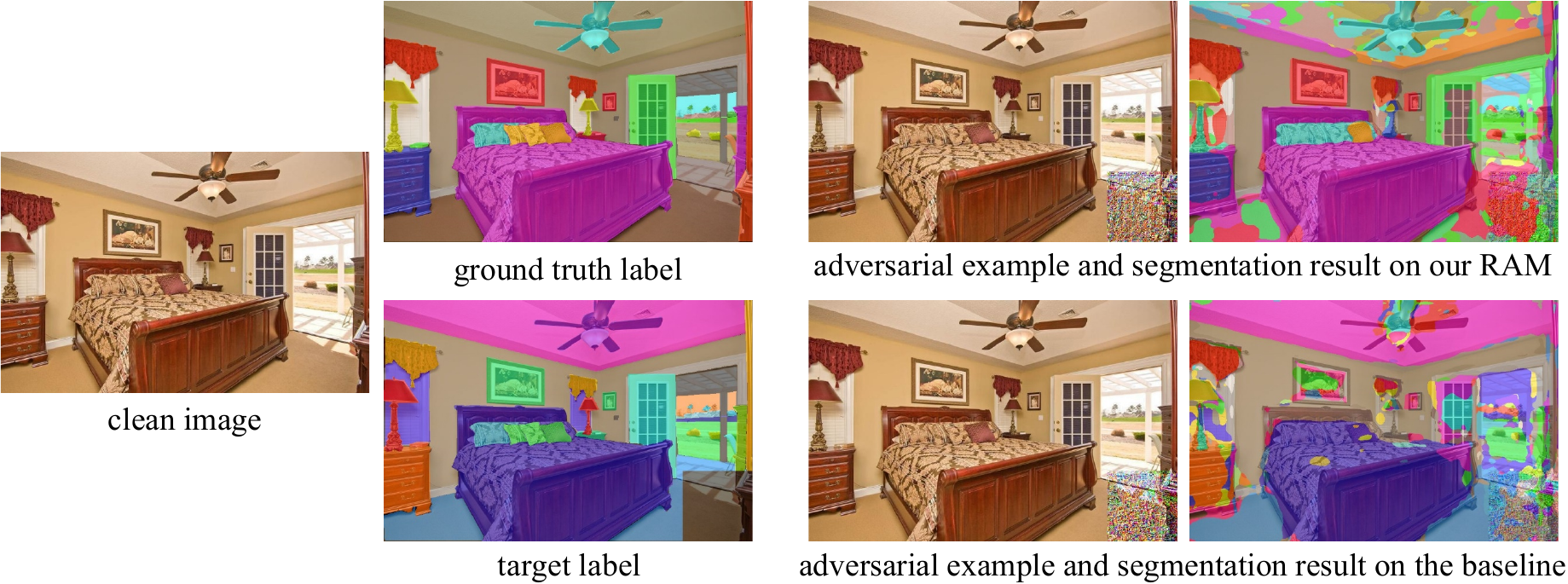}
    \caption{Visualization of the segmentation results of UPerNet~\cite{xiao2018unified}/DeiT~\cite{touvron2021training} model with our proposed RAM and the baseline under the attack setting of Permute.}
    \label{fig:visual_upernet}
    \end{subfigure}
    \begin{subfigure}[t]{\textwidth}
    \centering
    \includegraphics[width=\textwidth]{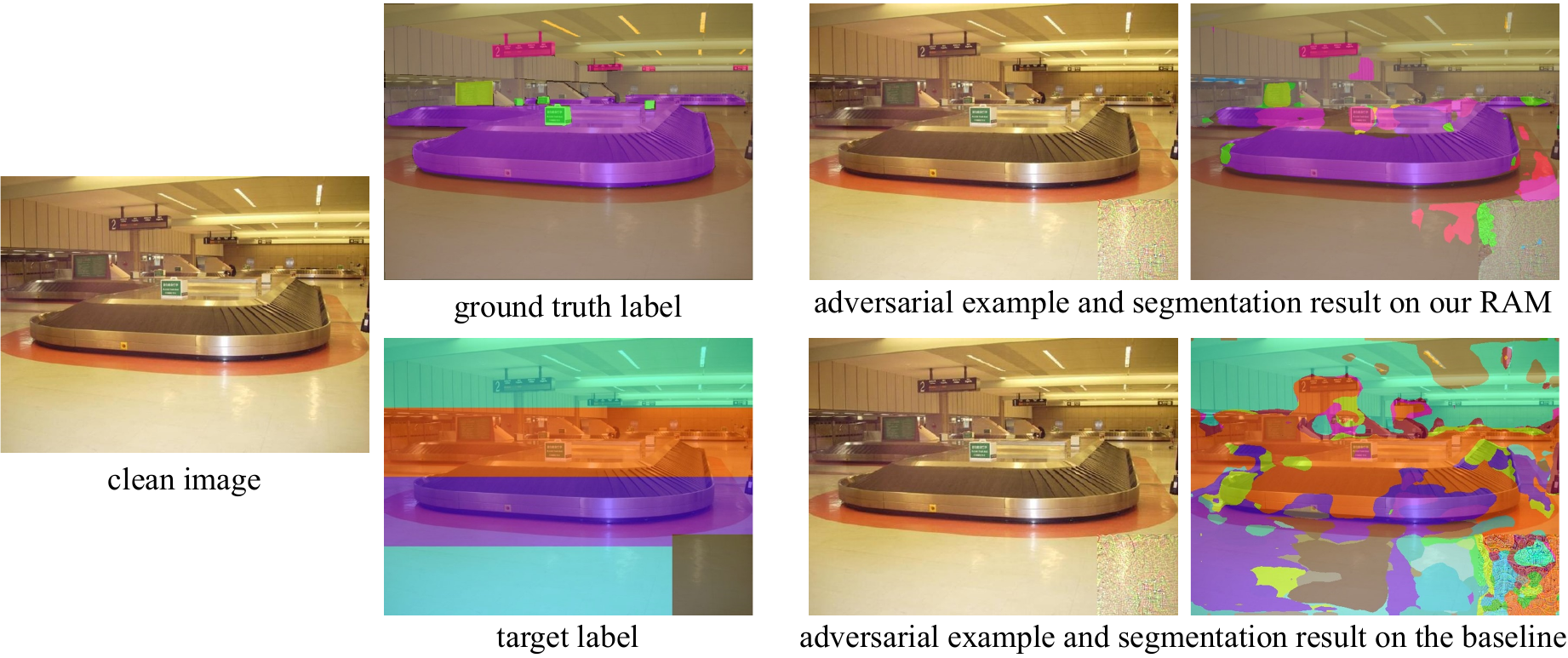}
    \caption{Visualization of the segmentation results of SegFormer~\cite{xie2021segformer}/MiT~\cite{xie2021segformer} model with our proposed RAM and the baseline under the attack setting of Strip.}
    \label{fig:visual_segformer}
  \end{subfigure}
  \caption{More visualizations about the comparison of segmentation results of different models.}
  \label{fig:more_visual}
\end{figure*}

\section{More Visualization Results}
\label{sec:visual}

  In this section, we visualize more segmentation results of semantic segmentation models under different attack scenarios in~\cref{fig:more_visual}.

  In~\cref{fig:visual_upernet}, we compare the segmentation results of UPerNet~\cite{xiao2018unified}/DeiT~\cite{touvron2021training} model with our proposed RAM and the baseline under the attack setting of Permute. When the baseline model is faced with an input image with the adversarial patch, the segmentation result almost completely meets the adversary's intention, that is, most areas of the segmentation results are consistent with the target labels determined by the adversary. As a comparison, when the model is with our proposed RAM, the adversarial patch less affects the output of the model, and most of the regions maintain correct segmentation results.

  In~\cref{fig:visual_segformer}, we compare the segmentation results of SegFormer~\cite{xie2021segformer}/MiT~\cite{xie2021segformer} model with our proposed RAM and the baseline under the attack setting of Strip. The segmentation results of the baseline model are very close to the strip pattern designed by the adversary, while the segmentation results of our RAM are rarely affected.

  Both of the visualization results demonstrate that the baseline model is quite vulnerable to patch-based attacks, while our RAM method significantly improves the robustness of semantic segmentation models against various attack scenarios.

\begin{table*}[!t]
  \centering
    \caption{The robustness of semantic segmentation models against two patch-based targeted attack settings. The results in the table are the mIoU between the segmentation results of adversarial images and the target labels determined by the adversary. The lower mIoU means the better robustness of the model. Details about the attack settings are illustrated in \cref{sec:setting}.}
    \begin{tabular}{c|ccc}
      \toprule
      \multirow{2}[2]{*}{Setting} & clean$\uparrow$ & Permute$\downarrow$ & Strip$\downarrow$ \\
            & (mIoU/\%) & (mIoU/\%) & (mIoU/\%) \\
      \midrule
      Baseline & 48.25  & 49.32  & 40.01  \\
      RAD+MAS & 48.16  & 45.61  & 35.17  \\
      MAS+RAD (ours) & \textbf{48.30} & \textbf{33.72} & \textbf{22.63} \\
      \bottomrule
    \end{tabular}%
  \label{tab:robustness}
\end{table*}

\begin{table*}[!t]
  \centering
    \caption{The robustness of semantic segmentation models against two patch-based targeted attack settings. The results in the table are the mIoU between the segmentation results of adversarial images and the target labels determined by the adversary. The lower mIoU means the better robustness of the model. Details about the attack settings are illustrated in \cref{sec:setting}.}
    \resizebox{\linewidth}{!}{
      \begin{tabular}{c|cc|cc|cc|cc}
        \toprule
        Model & \multicolumn{4}{c|}{Global Attention} & \multicolumn{4}{c}{Local Attention} \\
        \cmidrule{2-9}(Head/Backbone) & \multicolumn{2}{c|}{Nonlocal~\cite{wang2018nonlocal}/R50~\cite{he2016deep}} & \multicolumn{2}{c|}{SegFormer~\cite{xie2021segformer}/MiT~\cite{xie2021segformer}} & \multicolumn{2}{c|}{SeMask~\cite{jain2021semask}/Swin~\cite{liu2021swin}} & \multicolumn{2}{c}{UPerNet~\cite{xiao2018unified}/Swin~\cite{liu2021swin}} \\
        \midrule
              & clean$\uparrow$ & attack$\uparrow$ & clean$\uparrow$ & attack$\uparrow$ & clean$\uparrow$ & attack$\uparrow$ & clean$\uparrow$ & attack$\uparrow$ \\
              & (pAcc/\%) & (pAcc/\%) & (pAcc/\%) & (pAcc/\%) & (pAcc/\%) & (pAcc/\%) & (pAcc/\%) & (pAcc/\%) \\
        \midrule
        Baseline & 97.18  & 85.12  & 97.44  & 86.35  & 97.35  & 87.89  & 96.34  & 87.67  \\
        Certified~[1*] & 90.30  & 41.96  & 92.56  & 43.29  & 90.57  & 47.58  & 88.65  & 46.49  \\
        \midrule
        RAM (ours) & 96.87  & \textbf{87.96} & 97.13  & \textbf{88.34} & 96.34  & \textbf{90.37} & 95.79  & \textbf{89.53} \\
        \bottomrule
      \end{tabular}%
    }
  \label{tab:robustness}
\end{table*}

\end{appendices}


\end{document}